\documentclass[10pt,twocolumn,letterpaper]{article}

\usepackage{iccp}
\usepackage{times}
\usepackage{amsmath}
\usepackage{amssymb}
\usepackage{url}
\usepackage{graphicx}
\usepackage{color}
\usepackage{epstopdf}

\graphicspath{ {figures/} }

\newcommand{\Real}{\mathbb R}
\newcommand{\Natural}{\mathbb N}

\iccpfinalcopy 

\ificcpfinal\fi
\begin{document}

\title{Deep Convolutional Denoising of Low-Light Images}

\author{Tal Remez \textsuperscript{1}\\
	{\tt\small talremez@mail.tau.ac.il}
	\and
	Or Litany \textsuperscript{1}\\
	{\tt\small or.litany@gmail.com}
	\and
	Raja Giryes \textsuperscript{1}\\
	{\tt\small raja@tauex.tau.ac.il }
	\and
	Alex M. Bronstein \textsuperscript{2}\\
	{\tt\small bron@cs.technion.ac.il }
	\\\textsuperscript{1} School of Electrical Engineering, Tel-Aviv University, Israel
	\\\textsuperscript{2} Computer Science Department, Technion - IIT, Israel
}

\maketitle
\thispagestyle{empty}

\begin{abstract}
Poisson distribution is used for modeling noise in photon-limited imaging. While canonical examples include relatively exotic types of sensing like spectral imaging or astronomy, the problem is relevant to regular photography now more than ever due to the booming market for mobile cameras. Restricted form factor limits the amount of absorbed light, thus computational post-processing is called for. In this paper, we make use of the powerful framework of deep convolutional neural networks for Poisson denoising. We demonstrate how by training the same network with images having a specific peak value, our denoiser outperforms previous state-of-the-art by a large margin both visually and quantitatively. Being flexible and data-driven, our solution resolves the heavy ad hoc engineering used in previous methods and is an order of magnitude faster. We further show that by adding a reasonable prior on the class of the image being processed, another significant boost in performance is achieved.
\end{abstract}

\section{Introduction}
\begin{figure*}[t]
	\centering
	\begin{tabular}{c@{\hskip 0.01\textwidth}c@{\hskip 0.01\textwidth}c@{\hskip 0.01\textwidth}c}
		\includegraphics[width = 0.23\textwidth]{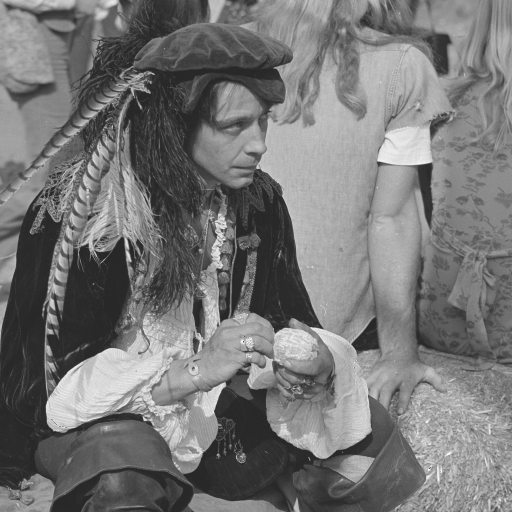} &
		\includegraphics[width = 0.23\textwidth]{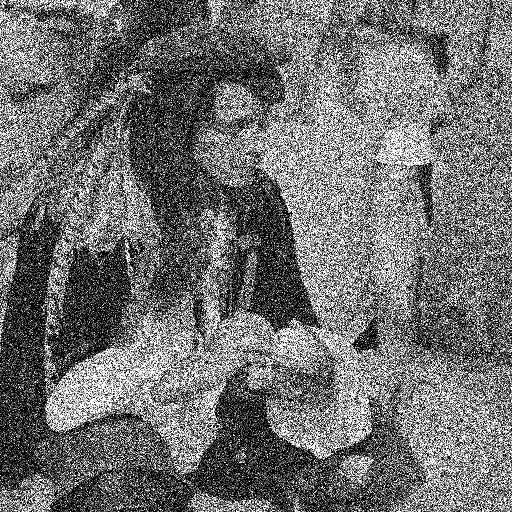} &
		\includegraphics[width = 0.23\textwidth]{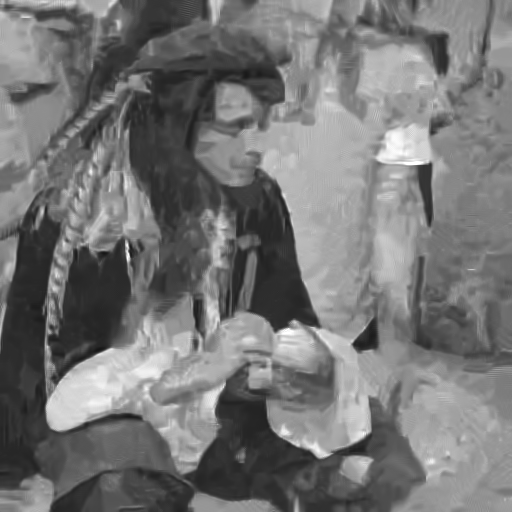} &
		\includegraphics[width = 0.23\textwidth]{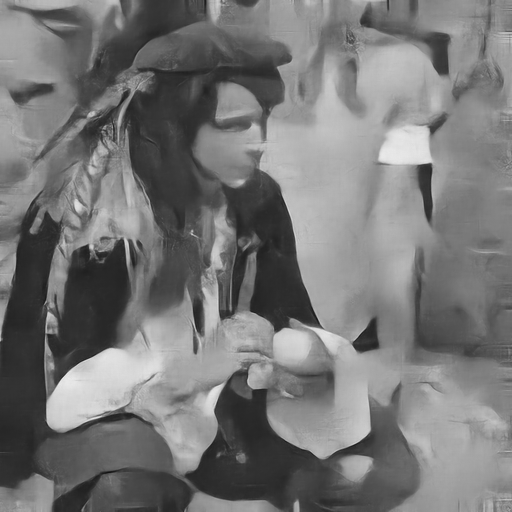} \\        
		Ground truth image& Noisy image& I+VST+BM3D \cite{Azzari16Variance} & Proposed DenoiseNet\\
		&  & $24.45$ dB & $24.77$ dB\\
	\end{tabular}      
	\smallskip 
	\label{teaser}
	\caption{\small \textbf{Perceptual comparison of proposed method.} The proposed denoiser produces visually more pleasant results and avoids artifacts commonly introduced by previous methods. The reader is encouraged to zoom-in to better appreciate the improvement. The presented image has a peak value of $4$. PSNR values are reported below the denoised images. }
\end{figure*}

Poisson noise, also known as shot noise, appears in many applications in various fields ranging from medical imaging to astronomy. It becomes dominant especially in the case of low photon count, as in the case of photography in low-light conditions. 
The problem becomes even more severe in modern mobile cameras that have a small form factor which reduces the amount of light that reaches the sensor. In view of the fact that today most of the photos are captured by smartphones, efficient techniques for Poisson noise removal are needed for improving the quality of images captured by these devices in low-light conditions. 

In the setup of Gaussian noise removal, which was studied much more than the Poisson counterpart but is less realistic especially in the low-light regime, the state-of-the-art performance is achieved using neural network-based strategies \cite{burger2012image, Chen16Trainable, vemulapalli2016deep}. Especially appealing are convolutional neural network (CNN) based solutions, which offer several advantages. First, they are known to have a high representation capability, thus, potentially enabling the learning of complex priors. In addition, they can be easily adapted to a certain data type merely by training on a specific dataset. Moreoever, being highly parallelizable, they lead in many cases to a fast computation on GPUs.

\paragraph{Our contribution.} In this paper we propose a novel fully convolutional residual neural network for Poisson noise removal. We demonstrate state-of-the-art results on several datasets compared to previous techniques. The improvements are noticeable both visually and quantitatively. The advantage of our proposed strategy stems from its simplicity. It does not rely on data models such as non-local similarity, sparse representation or Gaussian mixture models (GMM), which have been used in previous strategies \cite{dabov2007image, Giryes14Sparsity, Salmon13Poisson}, and only partially explain the structure of the data. By taking a supervised approach, and using the powerful representation capabilities demonstrated by deep neural networks, our method learns to remove the Poisson noise without explicitly relying on a model. Moreover, its convolutional structure makes it well suited to run on GPUs and other parallel hardware, thus, leading to running time that is of order of magnitude better than the other non convolutional model-based solutions \cite{dabov2007image, Giryes14Sparsity, Salmon13Poisson}.

The performance of our method is further improved when trained on a specific class of images. This scenario is particularly important as the vast majority of photos taken by mobile cameras contain a limited set of types of objects such as faces or landscapes. 


\section{The Poisson Denoising Problem}
\label{sec:background}
Let $X \in \left(\Natural \cup \left\{0 \right\}\right) ^{W \times H}$ denote a noisy image produced by a sensor. The goal of denoising is to recover a latent clean image $Y \in \Real_+^{W \times H}$ observed by the sensor.
In low-light imaging, the noise is dominated by shot noise; consequently, given the true value $Y_{ij}$ of the $(i,j)$-th pixel  expressed in number of photoelectrons, 
the corresponding value of the observed pixel $X_{ij}$ is an independent Poisson-distributed random variable with mean and variance $Y_{ij}$, i.e., $X_{ij} \sim \text{Poisson}(Y_{ij})$:
\begin{eqnarray}
	\label{eq:pois_dist}
	\text{P}(X_{ij} = n\big|Y_{ij} = \lambda) = \left\{ \begin{array}{cc}
		\frac{\lambda^{n}}{n!}e^{-\lambda} & \lambda >0 \\
		\delta_n & \lambda = 0.
	\end{array} \right.
\end{eqnarray}
Notice that Poisson noise is neither additive nor stationary, 
as its strength is dependent on the image intensity. Lower intensity in the image yields a stronger noise as
the SNR in each pixel is $\sqrt{Y_{ij}}$. Thus, it is natural to define the noise power in an image by the maximum value of $Y$ (its peak value). This is a good measure under the assumption that the intensity values are spread uniformly across the entire dynamic range, which holds for most natural images.

A very popular strategy \cite{Boulanger10Patch, Dupe09Proximal, Makitalo11Optimal, Zhang08Wavelets} for recovering $Y$ relies on variance-stabilizing transforms (VST),
such as Anscombe \cite{anscombe48transformation} and Fisz \cite{Fisz55Limiting},
that convert Poisson noise to be approximately white  Gaussian with unit variance. Thus, it is possible to solve the Poisson denoising problem using one of the numerous denoisers developed for Gaussian noise (e.g. \cite{burger2012image, Chen16Trainable, dabov2007image, Dong15Image, Dong13Nonlocally, Mairal09Non, Schmidt10Generative, vemulapalli2016deep}).

The problem with these approximations is the fact that they cease to hold for very low intensity values \cite{Makitalo11Optimal,Salmon13Poisson}. One strategy that has been used to overcome this deficiency correct the estimated varaince by applying the Gaussian denoiser and the Anscombe transform iteratively \cite{Azzari16Variance}. Another technique that also relies on Gaussian denoisers bypasses the need of using any type of VST by using the ``plug and play'' scheme \cite{Rond16Poisson}. 

An alternative approach is to develop new methods that are adapted to the Poisson noisy data directly \cite{ Danielyan11Deblurring, Deledalle10Poisson, Feng15Fast, Figueiredo10Restoration, Giryes14Sparsity, Rodrigo11Efficient, Remez15Picture, Salmon13Poisson,Willett03Platelets, Zhang12Novel}. For example, the work in \cite{Salmon13Poisson} proposed the non-local sparse PCA (NLSPCA) technique that relies on GMM \cite{Yu12Solving}. In \cite{Giryes14Sparsity},
a different strategy has been proposed, the sparse Poisson denoising algorithm (SPDA), which relies on sparse coding and dictionary learning. 
In \cite{Feng15Fast}, a nonlinear diffusion based neural network, previously proposed for Gaussian denosing \cite{Chen16Trainable}, has been adapted to the Poisson noise setting under the name of  trained reaction diffusion models for Poisson denoising (TRDPD).

A popular strategy improving the performance of many Poisson denoising algorithms is binning \cite{Salmon13Poisson}. Instead of processing the noisy image directly, a low-resolution version of the image with higher SNR is generated by aggregation of nearby pixels. Then, a given Poisson denoising technique is applied followed by simple linear interpolation to get back to the original high resolution image. The binning technique trades off spatial resolution and SNR, and has been shown to be useful in the very low SNR regimes.

\section{Denoising by DenoiseNet}
\label{sec:poissNet}

To recover the clean image $Y$ from its realization $X$, which is contaminated with Poisson noise, we propose a fully convolutional  neural network. Our architecture, denoted as DenoiseNet, is inspired by the network suggested in \cite{vemulapalli2016deep} for the purpose of super-resolution as the network estimates the difference between the noisy image and its clean counterpart. It also bares resemblance to the residual network introduced in \cite{He16Deep} since the weight gradients propagate to each layer both through its following layer and directly from the loss function. 

\subsection{Network architecture}
\begin{figure}[tb] 
	\includegraphics[width=1\linewidth]{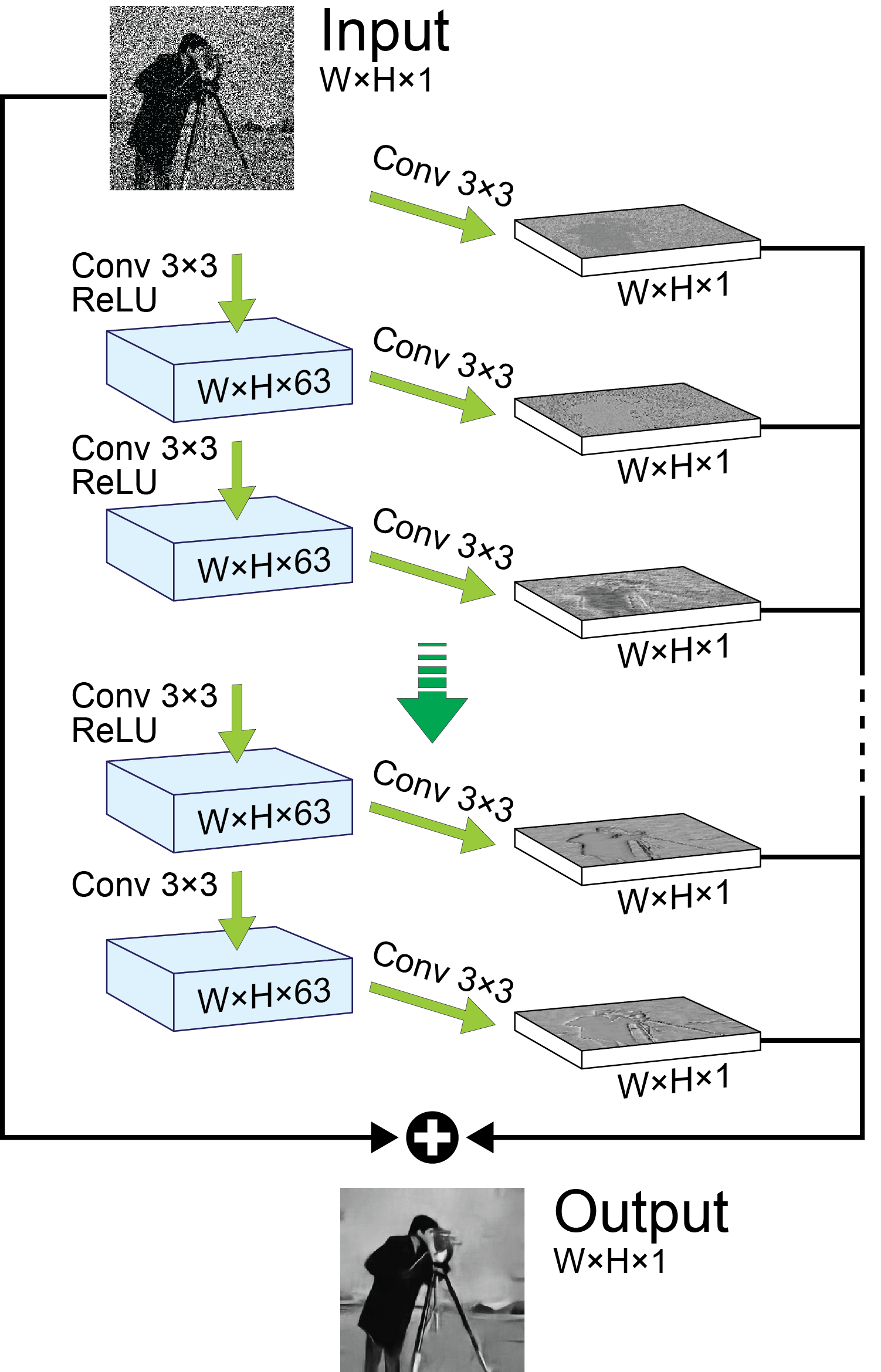}
	\caption{DenoiseNet architecture.}
	\label{fig_denoiseNet}
\end{figure}

The DenoiseNet architecture is shown in Figure ~\ref{fig_denoiseNet}.
The network receives a noisy grayscale image
as the input and produces an estimate of the original clean image.
At each layer, we convolve the previous layer output with $64$ kernels of size $3 \times 3$ using a stride of $1$. The first $63$ output channels are used for calculating the next steps, whereas the last channel is extracted to be directly combined with the input image to predict the clean output. Thus, these extracted layers can be viewed as negative noise components as their sum cancels out the noise. We build a deep network comprising $20$ convolutional layers, where the first $18$ use the \textit{ReLU} nonlinearity, while the last two are kept entirely linear. 

\subsection{Implementation details}
\label{sec_implementation_details}
We implemented the network in TensorFlow \cite{abadi2015tensorflow} and trained it for $120K$ iterations, which roughly took $72$ hours on a Titan-X GPU on a set of $8000$ images from the PASCAL VOC dataset \cite{everingham2010pascal}. We used mini-batches of $64$ patches of size $128
\times 128$. Images were converted to YCbCr and the Y channel was used as the input grayscale image after being scaled by the peak value and shifted by $-1/2$. For data augmenttion purposes, during training, image patches were randomly cropped from the training images and flipped about the vertical axis. Also, noise realization was randomzied.  Training was done using the ADAM optimizer \cite{DBLP:journals/corr/KingmaB14} with the learning rate $\alpha=10^{-4}$, $\beta_1=0.9$, $\beta_2=0.999$ and $\epsilon=10^{-8}$. Separate networks were trained respectively for different peak values. 
To avoid convolution artifacts at the borders of the patches, during training we used an $\ell_2$ loss on the central part of the patches cropping the outer $21$ pixels. At test time, images were padded with $21$ pixels using symmetric reflection before passing them through the network, and cropped back to their original size afterwards to give the final output.

\subsection{Class-aware denoising}
\label{sec:class_aware}
Having constructed a supervised framework for Poisson denoising, it is natural to seek for additional benefits of its inherent flexibility to fine-tune to specific data. One possibility to exploit this property, which we propose here, is to build class-specific denoisers, that is, to restrict the training data to a specific semantic class in order to boost the performance on it. This assumption is rather unrestrictive, as in many low-light imaging application the data being processed belong to a specific domain. In other settings, the class information can be provided manually by the user. For example, choosing face denoising for cleaning a personal photo collection. Alternatively, one could potentially train, yet, another deep network for automatic classification of the noisy images. 

The idea of combining classification with reconstruction has been previously proposed by 
\cite{baker2002limits}, which also dubbed it  \textit{recogstruction}. In their work, 
the authors set a bound on super-resolution performance and showed it can be broken when a face-prior is used. Several other studies have shown that it is beneficial to design a strategy for a specific class. For example, in \cite{Bryt08Compression} it has been shown that the design of a compression algorithm dedicated to faces improves over generic techniques targeting general images. Specifically for the class of faces, several face hallucination methods have been developed \cite{Wang14Comprehensive}, including face super-resolution and face sketch-photo synthesis techniques. 
In \cite{Joshi10Personal}, the authors showed that given a collection of photos of the same person it is possible to obtain a more faithful reconstruction of the face from a blury image.  In \cite{Iizuka16Let, zhang2016colorful}  class labeling at a pixel-level is used for the colorization of gray-scale images. In \cite{Anwar15Class}, the subspaces attenuated by blur kernels for specific classes are learned, thus improving the deblurring performance. 

Building on the success demonstrated in the aforementioned body of work, in this paper, we propose to use semantic classes as a prior and build class-aware denoisers. Different from previous methods, our model is made class-aware via training and not by design, hence  it can be automatically extended to any type and number of classes.

\section{Experiments}
\label{sec:exp}
We tested DenoiseNet performance and compared to other methods on the following series of experiments. Whenever code for another technique was not publicly available, we evaluated our method on the same test set and compared with the reported scores. As a first experiment, DenoiseNet was tested on the test set of the dataset it had been trained on, namely PASCAL VOC \cite{pascal-voc-2010}. We then applied it on a commonly used test set of $10$ images and compared against $8$ other methods for $5$ different peak values in the range $[1,30]$. To further appreciate the performance on different data we compared against \cite{Feng15Fast} on $68$ images from the Berkeley segmentation dataset \cite{MartinFTM01}. We proceeded with a short exploration that suggests that applying binning and the Anscombe transform, which are common techniques in the Poisson denoising literature, does not improve our network performance. Lastly, we fine tuned our network using specific classes of images from ImageNet to demonstrate the additional boost obtained in performance provided by such a prior.

\subsection{PASCAL VOC images}
In this experiment we tested our network on a set of $1000$ images from PASCAL VOC \cite{pascal-voc-2010}. 
A comparison with the recent iterative Poisson image denoising via VST method (I+VST+BM3D \cite{Azzari16Variance}), {which is considered to be the leading method for Poisson denoising to date,} shows an improvement in PSNR across all tested peak values between $1$ and $30$ of as much as $0.71$ dB (see Table \ref{tab_pascal_psnr}). A different network was trained to handle each of the peak values using images from PASCAL VOC as described in Section \ref{sec_implementation_details}. Interestingly, the gain achieved by our method decays as the peak values decrease. 

\begin{figure}[t] 
	\includegraphics[width=0.48\textwidth]{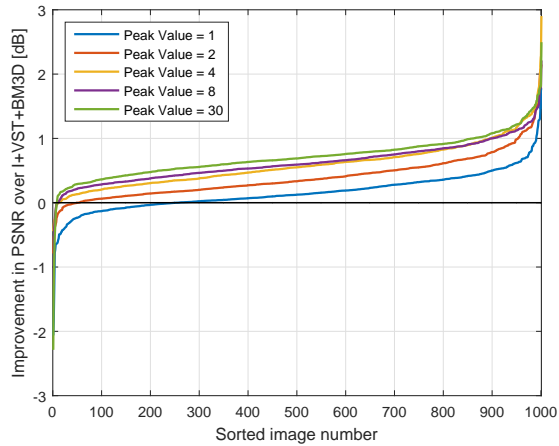}
	\caption{\textbf{Comparison of performance profile relative to I+VST+BM3D \cite{Azzari16Variance}.}  Image indices are sorted in ascending order of performance gain relative to I+VST+BM3D. The improvement of our method is demonstrated by (i) small zero-crossing point, and (ii) consistently higher PSNR values. The distribution reveals the statistical significance of the reported improvement. The comparison was made on images from PASCAL VOC and $15$ noise realizations per image.} 
	\label{pascal_s_curve}
\end{figure}

To examine the statistical significance of the improvement our method achieves, in Figure \ref{pascal_s_curve} we compare the gain in performance with respect to I+VST+BM3D achieved by our method. Image indices are sorted in ascending order of performance gain. A small zero-crossing value affirms our method outperforms I+VST+BM3D on the large majority of the images in the dataset. The plot visualizes the significant and consistent improvement in PSNR achieved by our method. An additional summary of the percentage of images from the dataset, on which each method outperformed the other is presented in Table \ref{tab_pascal_wins}. It is evident that with our method, a better reconstruction for peak values greater than $2$ is almost guaranteed. 
Denoising of several images from the test-set are visualized in Figure \ref{fig_large_pascal}. We encourage the reader to zoom-in to appreciate the significant improvement our method achieves, resulting in much more aesthetically pleasing images. 

\begin{table}[h]
	\centering
	\begin{tabular}{ l@{\hskip 0.01\textwidth}c@{\hskip 0.01\textwidth}c@{\hskip 0.01\textwidth}c@{\hskip 0.01\textwidth}c@{\hskip 0.01\textwidth}c@{\hskip 0.01\textwidth}c@{\hskip 0.01\textwidth}c@{\hskip 0.01\textwidth}c  }
		\hline \hline
		Peak 		& $1$ 	& $2$ 	& $4$ 	& $8$ 	& $30$ \\ \hline
		I+VST+BM3D  & $22.71$ & $23.70$ 	& $24.78$ 	& $26.08$ 	& $28.85$ \\
		DenoiseNet    & $\textbf{22.87}$ & $\textbf{24.09}$ 	& $\textbf{25.36}$ 	& $\textbf{26.70}$ 	& $\textbf{29.56}$ \\
		\hline
		PSNR gain    & $0.16$ & $0.39$ 	& $0.58$ 	& $0.62$ 	& $0.71$ \\       
		\hline\hline
	\end{tabular}  
	\vspace{2mm}
	\caption{\small \textbf{PSNR performance on PASCAL VOC \cite{pascal-voc-2010}.} Average PSNR values for different peak values on $1000$ test images and $15$ noise realizations per image. PSNR gain between the proposed method and I+VST+BM3D is presented in the bottom row.
	} 
	\label{tab_pascal_psnr}	
\end{table}

\begin{table}[h]
	\centering
	\begin{tabular}{ l@{\hskip 0.01\textwidth}c@{\hskip 0.01\textwidth}c@{\hskip 0.01\textwidth}c@{\hskip 0.01\textwidth}c@{\hskip 0.01\textwidth}c@{\hskip 0.01\textwidth}c@{\hskip 0.01\textwidth}c@{\hskip 0.01\textwidth}c  }
		\hline \hline
		Peak 		& $1$ 		& $2$ 		& $4$ 		& $8$ 		& $30$ 		\\ \hline
		I+VST+BM3D  & $26.0\%$	& $5.2\%$ 	& $1.1\%$ 	& $1\%$ 	& $0.8\%$ 	\\
		DenoiseNet    & $\textbf{74.0\%}$	& $\textbf{94.8\%}$ 	& $\textbf{98.9\%}$ 	& $\textbf{99\%}$ 	& $\textbf{99.2\%}$ 	\\
		\hline\hline
	\end{tabular}  
	\vspace{2mm}
	\caption{\small \textbf{Wins on PASCAL VOC \cite{pascal-voc-2010}.} Presented is the percentage of images out of the $1,000$ image test set on which each of the compared algorithms outperformed the others.  
	} 
	\label{tab_pascal_wins}	
\end{table}

\subsection{Standard image set}
In this experiment we evaluated our method on the standard set of images used by previous works. Evaluation was performed for peak values in the range of $1$ to $30$. PSNR values and running time presented in Table \ref{tab_classic_image_psnr} show that our method outperforms all other methods by a significant margin for the large majority of the images and almost all peak values. In addition, the execution time is orders of magnitude faster on a Titan-X GPU taking only $37$ milliseconds for a $256\times 256$ image, and is comparable to other methods when it runs on an{Intel E5-2630 $2.20$GHz CPU, taking $1.3$ seconds. A qualitative example can be seen in Figure \ref{teaser} showing the \textit{man} image denoised by DenoiseNet and by I+VST+BM3D\cite{Azzari16Variance} for a peak value of $4$.
	
	\begin{table*}[h]
		\centering
		\begin{tabular}{ l@{\hskip 0.01\textwidth}|c@{\hskip 0.01\textwidth}|c@{\hskip 0.01\textwidth}c@{\hskip 0.01\textwidth}c@{\hskip 0.01\textwidth}c@{\hskip 0.01\textwidth}c@{\hskip 0.01\textwidth}c@{\hskip 0.01\textwidth}c@{\hskip 0.01\textwidth}c@{\hskip 0.01\textwidth}c@{\hskip 0.01\textwidth}c@{\hskip 0.01\textwidth}|l@{\hskip 0.01\textwidth}  }
			\hline \hline
			Method &  Peak &  Flag &  House &  Cam &  Man &  Bridge &  Saturn &  Peppers &  Boat &  Couple &  Hill &  Time \\
			\hline
			\hline
			
			NLSPCA &  	& $19.68$ & $21.57$ & $20.25$ & $21.46$ & $19.02$ & $24.75$ & $19.5$ & $21.19$ & $21.14$ & $21.94$ & $86$s \\
			
			NLSPCA bin &	& $15.77$ & $20.78$ & $18.4$ & $19.87$ 	& $18.26$ & $22.83$ & $17.78$ & $20.19$ & $20.11$ & $20.82$ & $16$s \\
			
			SPDA &			& $\textbf{22.97}$ & $22.14$ & $20.15$ & - 		& $19.30$  & $27.05$ & $19.97$ & - 		& - 		& - 	& $5$h \\
			
			SPDA bin &		& $18.99$ & $20.99$ & $19.43$ & $21.15$ & $18.84$ & $27.40$ & $18.93$ & $21.19$ & $20.97$ & $21.5$ & $25$min \\
			
			P4IP &		$1$	& $19.07$ & $22.67$ & $20.54$ & - 		& $19.31$ & $27.05$ & $20.07$ & - 		& - 		& -	 	& few mins \\
			
			VST+BM3D &		& $18.46$ & $21.64$ & $20.19$ & $21.62$ & $19.43$ & $25.82$ & $19.71$ & $21.47$ & $21.14$ & $21.92$ & $0.78$s \\
			
			VST+BM3D bin &	& $19.28$ & $22.53$ & $20.69$ & $22.07$ & $19.59$ & $\textbf{27.59}$ & $20.22$ & $21.97$ & $21.81$ & $22.72$ & $0.10$s \\
			
			I+VST+BM3D &	& $19.74$ & $\textbf{23.04}$ & $21.07$ & $22.30$ & $\textbf{19.86}$  & $27.27$ & $20.44$ & $22.17$ & $22.08$ & $\textbf{22.85}$ & $0.82$s \\
			
			DenoiseNet &		& $19.45$ & $22.87$ & $\textbf{21.59}$ & $\textbf{22.49}$ & $19.83$ & $26.26$ & $\textbf{21.43}$ & $\textbf{22.38}$ & $\textbf{22.11}$ & $22.82$ & $0.04$s/$1.3$s \\
			\hline
			
			NLSPCA & 		& $19.70$ & $23.16$ & $20.64$ & $22.37$ & $19.43$ & $26.88$ & $20.48$ & $21.83$ & $21.75$ & $22.68$ & $87$s \\
			
			NLSPCA bin &	& $15.52$ & $20.85$ & $18.35$ & $19.87$ & $18.32$ & $21.27$ & $17.78$ & $20.29$ & $20.21$ & $20.98$ & $12$s \\
			
			SPDA &			& $\textbf{24.72}$ & $24.37$ & $21.35$ & - 		& $20.17$ & $29.13$ & $21.18$ & - 		& - 	  & - 		& $6h$ \\
			
			SPDA bin &		& $19.26$ & $21.12$ & $19.53$ & $21.66$ & $18.87$ & $28.54$ & $19.17$ & $21.43$ & $21.24$ & $21.94$ & $25$min \\
			
			P4IP &	$2$ 	& $21.04$ & $24.65$ & $21.87$ & - 		& $20.16$ & $28.93$ & $21.33$ & - & - & - & few mins \\
			
			VST+BM3D &		& $20.79$ & $23.79$ & $21.97$ & $23.11$ & $20.49$ & $27.95$ & $22.02$ & $22.90$ & $22.65$ & $23.34$ & $0.82$s \\
			
			VST+BM3D bin && $19.91$ & $24.10$ & $21.43$ & $23.03$ & $20.36$ & $\textbf{29.26}$ & $21.45$ & $22.92$ & $22.84$ & $23.75$ & $0.10$s \\
			
			I+VST+BM3D &	& $21.18$ & $24.62$ & $22.25$ & $23.40$ & $20.69$ & $28.85$ & $21.93$ & $23.30$ & $23.12$ & $23.88$ & $0.82$s \\
			
			DenoiseNet &		& $21.38$ & $\textbf{24.77}$ & $\textbf{23.25}$ & $\textbf{23.64}$ & $\textbf{20.80}$ & $28.37$ & $\textbf{23.19}$ & $\textbf{23.66}$ & $\textbf{23.30}$ & $\textbf{23.95}$ & $0.04$s/$1.3$s \\
			\hline
			
			NLSPCA & 	 	& $20.15$ & $24.26$ & $20.97$ & $22.93$ & $20.21$ & $27.99$ & $21.07$ & $22.49$ & $22.33$ & $23.51$ & $123$s \\
			
			NLSPCA bin &	& $15.52$ & $20.94$ & $18.27$ & $19.88$ & $18.32$ & $22.02$ & $17.72$ & $20.29$ & $20.25$ & $20.99$ & $13$s \\
			
			SPDA &			& $\textbf{25.76}$ & $25.3$ & $21.72$ & -& $20.53$ & $\textbf{31.13}$ & $22.2$ & - & -& - & $8$h\\
			
			SPDA bin &		& $19.42$ & $22.07$ & $19.95$ & $22.18$ & $19.26$ & $29.71$ & $20.19$ & $21.76$ & $21.69$ & $22.82$ & $31$min \\
			
			P4IP &		$4$	& $22.49$ & $26.33$ & $23.29$ & $24.66$ & $21.11$ & $30.82$ & $23.88$ & $24.10$ & $23.99$ & $25.28$ & few mins \\
			
			VST+BM3D &		& $22.93$ & $25.49$ & $23.82$ & $24.32$ & $21.51$ & $29.41$ & $24.01$ & $24.16$ & $24.10$ & $24.47$ & $0.74$s \\
			
			VST+BM3D bin &	& $20.43$ & $25.49$ & $22.22$ & $23.99$ & $21.13$ & $30.87$ & $22.57$ & $23.92$ & $23.84$ & $24.69$ & $0.10$s \\
			
			I+VST+BM3D &	& $23.51$ & $26.07$ & $24.10$ & $24.52$ & $21.71$ & $30.38$ & $24.04$ & $24.53$ & $24.34$ & $24.82$ & $1.41$s \\
			
			DenoiseNet &		& $23.18$ & $\textbf{26.59}$ & $\textbf{24.87}$ & $\textbf{24.77}$ & $\textbf{21.81}$ & $30.02$ & $\textbf{24.83}$ & $\textbf{24.86}$ & $\textbf{24.60}$ & $\textbf{25.01}$ & $0.04$s/$1.3$s \\
			\hline
			
			NLSPCA & 	 	& $14.87$ & $20.87$ & $18.21$ & $19.76$ & $18.23$ & $21.44$ & $17.67$ & $20.20$ & $20.21$ & $20.93$ & $60$s \\
			
			SPDA &			& $\textbf{26.85}$ & $26.36$ & $22.24$ & $24.36$ & $21.05$ & $32.39$ & $22.89$ & $23.50$ & $23.37$ & $24.93$ & days \\ 
			
			P4IP &		$8$	& $23.10$ & $27.36$ & $24.49$ & $24.96$ & $21.68$ & $\textbf{32.88}$ & $24.94$ & $25.03$ & $25.06$ & $24.50$ & $167$s \\
			
			I+VST+BM3D &	& $25.54$ & $27.95$ & $25.74$ & $25.81$ & $22.72$ & $32.35$ & $25.90$ & $25.95$ & $25.79$ & $26.06$ & $5.1$s \\ 
			
			DenoiseNet &		& $25.73$ & $\textbf{28.42}$ & $\textbf{26.35}$ & $\textbf{26.10}$ & $\textbf{22.91}$ & $32.28$ & $\textbf{26.45}$ & $\textbf{26.23}$ & $\textbf{26.11}$ & $\textbf{26.26}$ & $0.04$s/$1.3$s \\
			\hline 
			
			NLSPCA & 		&  $14.78$ & $18.83$ & $17.98$ & $19.39$ & $18.03$ & $21.41$ & $17.06$ & $19.92$ & $19.98$ & $20.60$ & $92$s \\
			
			SPDA &			& $27.10$ & $27.06$ & $22.47$ & $25.02$	& $21.22$ & $35.08$ & $23.61$ & $24.55$ & $24.06$ & $25.88$ & days \\ 
			
			P4IP &	 $30$	& $27.02$ & $29.85$ & $27.28$ & $26.52$ & $23.07$ & $36.03$ & $27.33$ & $26.98$ & $27.22$ & $27.01$ & $149$s\\
			
			I+VST+BM3D &	& $\textbf{29.09}$ & $31.35$ & $28.55$ & $28.37$ & $25.08$ & $36.03$ & $29.08$ & $28.79$ & $28.80$ & $28.62$ & $4.5$s \\
			
			DenoiseNet &		& $28.94$ & $\textbf{31.67}$ & $\textbf{29.21}$ & $\textbf{28.74}$ & $\textbf{25.42}$ & $\textbf{36.20}$ & $\textbf{29.77}$ & $\textbf{29.06}$ & $\textbf{29.13}$ & $\textbf{28.71}$ & $0.04$s/$1.3$s \\
			
			\hline\hline 
			
		\end{tabular}  
		\vspace{2mm}
		\caption{\small \textbf{Performance on standard images.} Numeric values represent PSNR in dB averaged over five noise realizations. Values for prior art algorithms for peak values of $1-4$ were taken from \cite{Azzari16Variance}. For the rest of the peak values we ran the code published by the authors; in the absence of optimal parameter settings, we used those for  peak$=4$. Timing values presented are averages for images of size $256\times256$, for DenoiseNet we present the run-time on GPU/CPU. 
		} 
		\label{tab_classic_image_psnr}	
	\end{table*}
	
	\subsection{Berkeley segmentation dataset}
	\label{sec_berkeley}
	In this experiment we tested our method on $68$ test images from the Berkeley dataset \cite{MartinFTM01} (as selected by \cite{roth2009fields}), and compared it with \cite{Feng15Fast} and I+VST+BM3D \cite{Azzari16Variance}. Note that we did not fine-tune our network to fit this dataset but rather used it after it had been trained on PASCAL VOC. Results are summarized in Table \ref{tab_68_psnr}. 
	The superiority of DenoiseNet over other methods is evident across all peak values. 
	Especially interesting is the improvement compared to \cite{Feng15Fast}, where a trainable nonlinear reaction diffusion network was tuned for Poisson denoising on this data. This suggests that a flexible network architecture may sometimes be preferable to a model-driven one, and especially in the case where training data are practically unlimited. Also note that both networks have similar run time. 
	
	\begin{table}[h]
		\centering
		\begin{tabular}{ l@{\hskip 0.01\textwidth}c@{\hskip 0.01\textwidth}c@{\hskip 0.01\textwidth}c@{\hskip 0.01\textwidth}c@{\hskip 0.01\textwidth}c@{\hskip 0.01\textwidth}c@{\hskip 0.01\textwidth}c  }
			\hline \hline
			Peak 				 & $1$ 				& $2$ 				& $4$ 				& $8$ 				\\ \hline
			NLSPCA 				 & $20.90$			& $21.60$			& $22.09$			& $22.38$			\\
			NLSPCA bin  		 & $19.89$			& $19.95$	 		& $19.95$			& $19.91$			\\
			VST+BM3D 			 & $21.01$			& $22.21$			& $23.54$			& $24.84$			\\
			VST+BM3D bin 		 & $21.39$			& $22.14$			& $22.87$			& $23.53$			\\
			I+VST+BM3D 		 	 & $21.66$			& $22.59$			& $23.69$			& $24.93$			\\
			TRDPD$^8_{5\times5}$ & $21.49$			& $22.54$			& $23.70$			& $24.96$			\\
			TRDPD$^8_{7\times7}$ & $21.60$			& $22.62$			& $23.84$			& $25.14$	 		\\
			DenoiseNet    		 & $\textbf{21.79}$ & $\textbf{22.90}$ 	& $\textbf{23.99}$ 	& $\textbf{25.30}$ 	\\ 
			\hline\hline
		\end{tabular}  
		\vspace{2mm}
		\caption{\small \textbf{PSNR performance $68$ images set from \cite{roth2009fields}.} Average PSNR values for different peak values on $68$ image test set from \cite{roth2009fields}. Results reported in \cite{Feng15Fast} were copied as is, and values for our method and for I+VST+BM3D \cite{Azzari16Variance} were added (averaging over $15$ noise realizations per image). Our network was trained on PASCAL VOC images as described in Section~\ref{sec_implementation_details}.
		} 
		\label{tab_68_psnr}	
	\end{table}

	\begin{figure*}[]
		\centering   
		\small
		\begin{tabular}{c@{\hskip 0.02\textwidth}c@{\hskip 0.02\textwidth}c@{\hskip 0.02\textwidth}c}    
			Ground Truth & I+VST+BM3D \cite{Azzari16Variance} & DenoiseNet & Class-specific DenoiseNet\\
			\includegraphics[width = 0.2\textwidth]{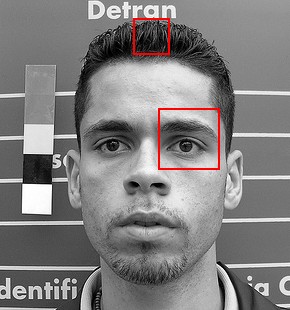} &
			\includegraphics[width = 0.2\textwidth]{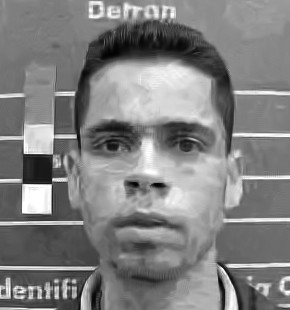} &
			\includegraphics[width = 0.2\textwidth]{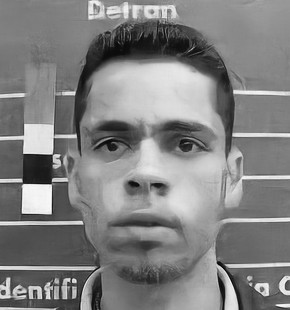}  &
			\includegraphics[width = 0.2\textwidth]{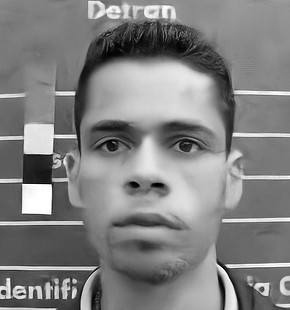} \\  
		\end{tabular}  
		\begin{tabular}{c@{\hskip 0.01\textwidth}c@{\hskip 0.02\textwidth}c@{\hskip 0.01\textwidth}c@{\hskip 0.02\textwidth}c@{\hskip 0.01\textwidth}c@{\hskip 0.02\textwidth}c@{\hskip 0.01\textwidth}c}
			
			\includegraphics[width = 0.095\textwidth]{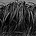} &
			\includegraphics[width = 0.095\textwidth]{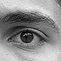} &
			\includegraphics[width = 0.095\textwidth]{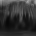} &
			\includegraphics[width = 0.095\textwidth]{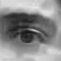} &
			\includegraphics[width = 0.095\textwidth]{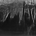} &
			\includegraphics[width = 0.095\textwidth]{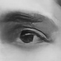} &
			\includegraphics[width = 0.095\textwidth]{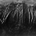} &
			\includegraphics[width = 0.095\textwidth]{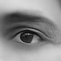} \\
			
			\multicolumn{2}{c}{\smallskip} &
			\multicolumn{2}{c}{$25.85$ dB} & 
			\multicolumn{2}{c}{$26.79$ dB} & 
			\multicolumn{2}{c}{$26.96$ dB} \\
		\end{tabular}  
		
		\begin{tabular}{c@{\hskip 0.02\textwidth}c@{\hskip 0.02\textwidth}c@{\hskip 0.02\textwidth}c}   
			
			\includegraphics[width = 0.2\textwidth]{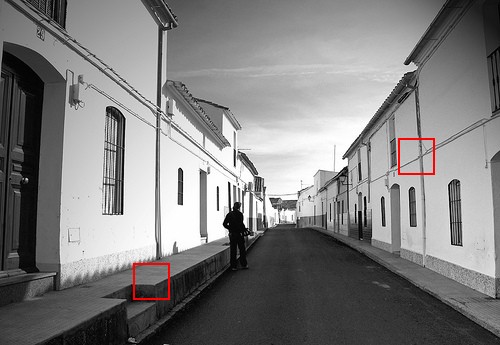} &
			\includegraphics[width = 0.2\textwidth]{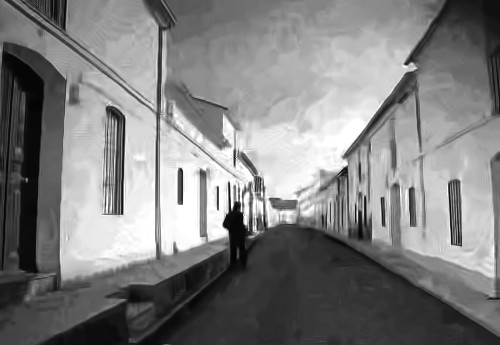} &
			\includegraphics[width = 0.2\textwidth]{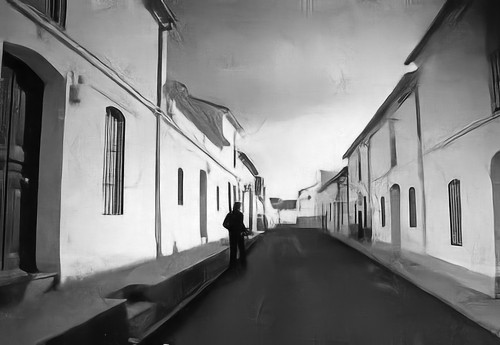}  &
			\includegraphics[width = 0.2\textwidth]{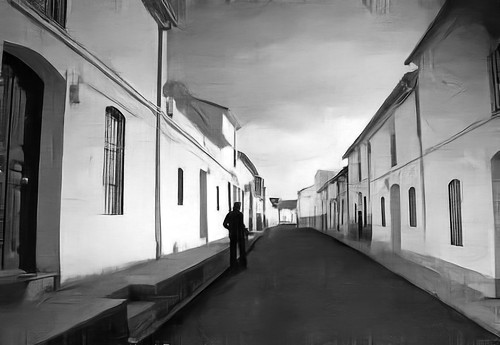} \\ 
		\end{tabular}  
		\begin{tabular}{c@{\hskip 0.01\textwidth}c@{\hskip 0.02\textwidth}c@{\hskip 0.01\textwidth}c@{\hskip 0.02\textwidth}c@{\hskip 0.01\textwidth}c@{\hskip 0.02\textwidth}c@{\hskip 0.01\textwidth}c}
			
			\includegraphics[width = 0.095\textwidth]{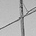} &
			\includegraphics[width = 0.095\textwidth]{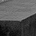} &
			\includegraphics[width = 0.095\textwidth]{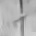} &
			\includegraphics[width = 0.095\textwidth]{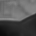} &
			\includegraphics[width = 0.095\textwidth]{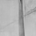} &
			\includegraphics[width = 0.095\textwidth]{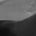} &
			\includegraphics[width = 0.095\textwidth]{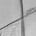} &
			\includegraphics[width = 0.095\textwidth]{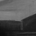} \\
			
			\multicolumn{2}{c}{\smallskip} &
			\multicolumn{2}{c}{$25.56$ dB} & 
			\multicolumn{2}{c}{$26.47$ dB} & 
			\multicolumn{2}{c}{$26.75$ dB} \\
		\end{tabular}  
		
		\begin{tabular}{c@{\hskip 0.02\textwidth}c@{\hskip 0.02\textwidth}c@{\hskip 0.02\textwidth}c}  
			\includegraphics[width = 0.2\textwidth]{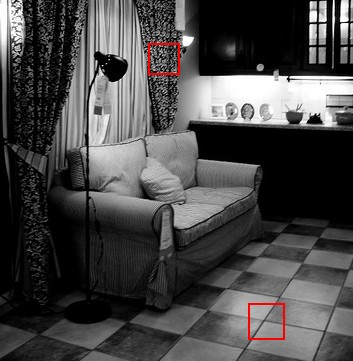} &
			\includegraphics[width = 0.2\textwidth]{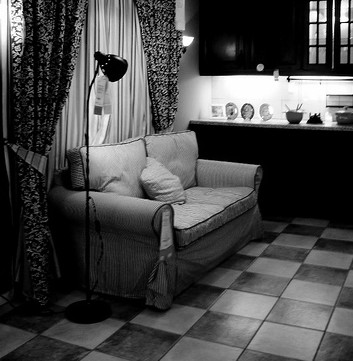} &
			\includegraphics[width = 0.2\textwidth]{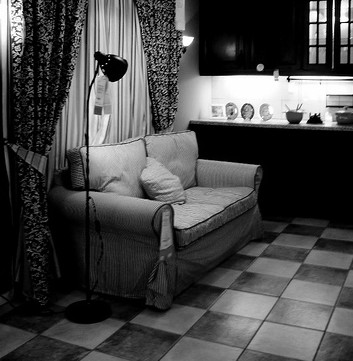}  &
			\includegraphics[width = 0.2\textwidth]{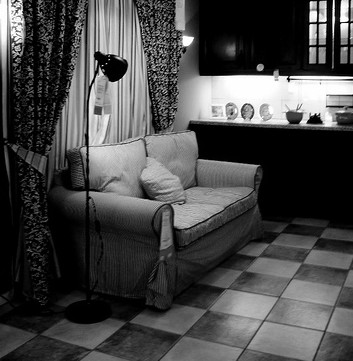}\\ 
		\end{tabular}  
		\begin{tabular}{c@{\hskip 0.01\textwidth}c@{\hskip 0.02\textwidth}c@{\hskip 0.01\textwidth}c@{\hskip 0.02\textwidth}c@{\hskip 0.01\textwidth}c@{\hskip 0.02\textwidth}c@{\hskip 0.01\textwidth}c}
			
			\includegraphics[width = 0.095\textwidth]{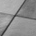} &
			\includegraphics[width = 0.095\textwidth]{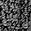} &
			\includegraphics[width = 0.095\textwidth]{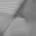} &
			\includegraphics[width = 0.095\textwidth]{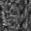} &
			\includegraphics[width = 0.095\textwidth]{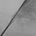} &
			\includegraphics[width = 0.095\textwidth]{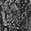} &
			\includegraphics[width = 0.095\textwidth]{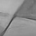} &
			\includegraphics[width = 0.095\textwidth]{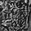} \\
			
			\multicolumn{2}{c}{\smallskip} &
			\multicolumn{2}{c}{$24.09$ dB} & 
			\multicolumn{2}{c}{$24.96$ dB} & 
			\multicolumn{2}{c}{$25.36$ dB} \\
		\end{tabular}  
		
		\begin{tabular}{c@{\hskip 0.02\textwidth}c@{\hskip 0.02\textwidth}c@{\hskip 0.02\textwidth}c}  
			\includegraphics[width = 0.2\textwidth]{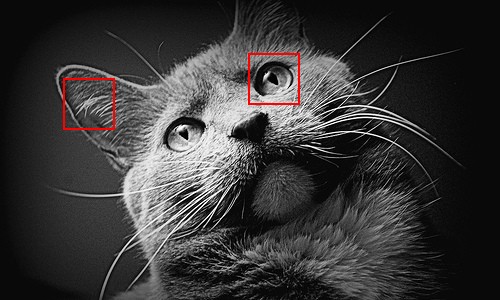} &
			\includegraphics[width = 0.2\textwidth]{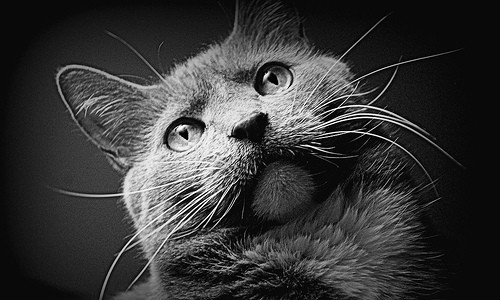} &
			\includegraphics[width = 0.2\textwidth]{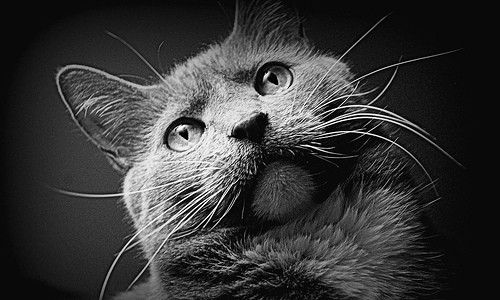}  &
			\includegraphics[width = 0.2\textwidth]{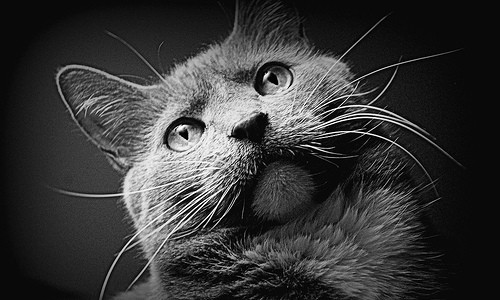}  \\ 
		\end{tabular}  
		\begin{tabular}{c@{\hskip 0.01\textwidth}c@{\hskip 0.02\textwidth}c@{\hskip 0.01\textwidth}c@{\hskip 0.02\textwidth}c@{\hskip 0.01\textwidth}c@{\hskip 0.02\textwidth}c@{\hskip 0.01\textwidth}c}
			
			\includegraphics[width = 0.095\textwidth]{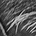} &
			\includegraphics[width = 0.095\textwidth]{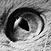} &
			\includegraphics[width = 0.095\textwidth]{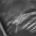} &
			\includegraphics[width = 0.095\textwidth]{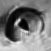} &
			\includegraphics[width = 0.095\textwidth]{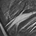} &
			\includegraphics[width = 0.095\textwidth]{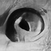} &
			\includegraphics[width = 0.095\textwidth]{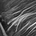} &
			\includegraphics[width = 0.095\textwidth]{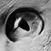} \\
			
			\multicolumn{2}{c}{\smallskip} &
			\multicolumn{2}{c}{$23.53$ dB} & 
			\multicolumn{2}{c}{$24.60$ dB} & 
			\multicolumn{2}{c}{$24.78$ dB} \\
		\end{tabular}   
		\vspace{-3mm}
		\caption{\textbf{Denoising examples from ImageNet.} Presented are images from ImageNet denoised by I+VST+BM3D and our class-agnostic and class-specific denoisers for peak $8$. PSNR values appear below each of the images.}
		\label{fig_large_imagenet}
	\end{figure*}

	\subsection{The influence of binning and VST}
	Two techniques have been shown in the past to boost reconstruction quality in Poisson denoising: the use of the Anscombe transform (and its inverse), and binning. To check whether these could also improve our proposed network we changed the architecture in the following way. We added a first layer with $4$ channels consisting of constant valued square kernels of sizes $n \times n$, where $n = 1, 3, 5, 7$. This has the effect of binning. This layer's weights were kept fixed at training phase. We further added a second layer realizing the non-linear Anscombe transform. The input was also transformed before adding it to the intermediate residual outputs. Finally, we added the exact inverse transform \cite{Makitalo11Optimal} after summing the residuals and the transformed noisy input image. Interestingly, these modifications had a negligible effect on the network performance. As for the reasons why or whether there would have been an effect for shallower networks, we leave these for future work. 
	
	\subsection{Class-aware denoising}
	In order to demonstrate the benefits of having a flexible architecture when an adaptation to a specific data type is required, we selected the following semantic classes: face, flower, street, living-room, and pet. About $1200$ images were collected for each of the $5$ semantic classes from ImageNet \cite{ImageNet15}.  Starting from the network that has been trained on PASCAL for peak $8$, we fine-tuned a separate network to each of the classes, thus, making them ''class-specific'' (as opposed to being ''class-agnostic'' before the fine-tunning process). Tuning procedure consisted of $45 \times 10^3$ training iterations with the same parameters used for the initial training. The images of each class were slip into training ($60\%$), validation ($20\%$) and test ($20\%$) sets.

	The performance of our class-specific networks compared to the class-agnostic baseline and I+VST+BM3D is summarized in Table \ref{tab_class_aware_psnr}. While the class-agnostic DenoiseNet outperforms I+VST+BM3D by as much as $0.6$ dB, its class-specific version boosts performance by additional $0.15$ to $0.31$ dB.
	A visual inspection is presented in Figure~\ref{fig_large_imagenet}, which demonstrates an already large improvement of our proposed class-agnostic method compared to previous methods, and yet an additional non-negligible boost attained by the class-aware network. We encourage the reader to zoom in, e.g., on the person's face, the living-room floor and curtains, the streets curbs and building facades, and the cat's eyes and fur to fully appreciate the improvement in visual quality. 
	Lastly, in Figure \ref{fig_confusion} we present a confusion matrix calculated for applying different denoiser class on all different image classes. It can be seen that the network has indeed learned distinguishable features per class and has specialized on it yielding higher PSNR with respect to all other class-aware denoisers for the majority of images.
	
	\begin{figure}[] 
		\includegraphics[width=0.47\textwidth]{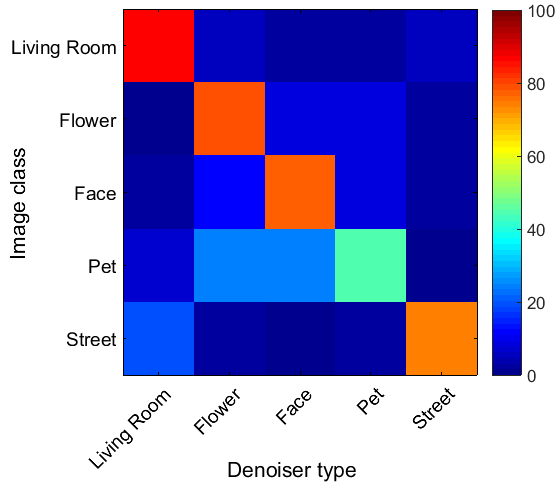}
		\caption{\textbf{Denoiser performance per semantic class.} Each row represents a specific semantic class of images while class-aware denoisers are represented as columns. The $(i,j)$-th element in the confusion matrix shows the probability of the $j$-th class-aware denoiser to outperform all other denoisers on the $i$-th class of images. The diagonal dominant structure indicates that each  denoiser specializes on a particular class.}
		\label{fig_confusion}
	\end{figure}
	
	\begin{table}[h]
		\centering
		\begin{tabular}{ l@{\hskip 0.01\textwidth}c@{\hskip 0.01\textwidth}c@{\hskip 0.01\textwidth}c@{\hskip 0.01\textwidth}c@{\hskip 0.01\textwidth}c@{\hskip 0.01\textwidth}c@{\hskip 0.01\textwidth}c@{\hskip 0.01\textwidth}c  }
			\hline \hline

			Image class 	& Face 				& Flower 			& Livingroom 		& Pet 			& Street 	\\ \hline
			I+VST+BM3D  	& $27.17$			& $26.16$ 			& $26.39$ 	 		& $25.95$ 		& $24.13$ 	\\
			Class-unaware   & $27.70$			& $26.68$ 			& $26.99$ 	 		& $26.39$ 		& $24.69$ 	\\
			Class-specific	& $\textbf{28.01}$	& $\textbf{26.93}$ 	& $\textbf{27.19}$ 	& $\textbf{26.54}$ 	&$\textbf{24.85}$ 	\\
			\hline\hline
		\end{tabular}  
		\vspace{2mm}
		\caption{\small \textbf{ Class-aware denoising on ImageNet data.} Presented is the average PSNR performance on specific class images from ImageNet. It is evident that DenoiseNet significantly outperforms I+VST+BM3D \cite{Azzari16Variance} even when it is class-agnostic by as much as $0.6$ dB, while the class-specific approach boosts performance by an additional $0.15$ to $0.31$ dB. We averaged $15$ noise realizations per image and $300$ images per class.
		} 
		\label{tab_class_aware_psnr}	
	\end{table}

	\subsection{``Under the hood'' of DenoiseNet}
	This section presents a few examples that we believe give insights about the noise estimation of DenoiseNet. In Figure \ref{fig_layer_selection} we show several denoised images with peak value $8$ and the error after $5, 10, $ and $20$ layers (rows $4-6$). Surprisingly, even though it has not been explicitly enforced at training, the error monotonically decreases with the layers' depth (see plots in row $7$ in Figure \ref{fig_layer_selection}). This non-trivial behavior is consistently produced by the network on almost all test images. To visualize which of the layers was the most dominant in the denoising process, we assign a different color to each layer and color each pixel according to the layer in which its value changed the most. The resulting image is shown in the bottom row of Figure \ref{fig_layer_selection}. It can be observed that the first few layers govern the majority of the pixels while the following ones mainly focus on recovering and enhancing the edges and textures that might have been degraded by the first layers.

\begin{figure*}[]
	\begin{centering}
		\begin{tabular}{c@{\hskip 0.007\textwidth}c@{\hskip 0.005\textwidth}c@{\hskip 0.005\textwidth}c@{\hskip 0.005\textwidth}c@{\hskip 0.005\textwidth}c@{\hskip 0.005\textwidth}c}
			\hspace{-1mm}\parbox[b][4em][s]{0.16\textwidth}{Ground truth}&
			\includegraphics[height = 0.15\textwidth]{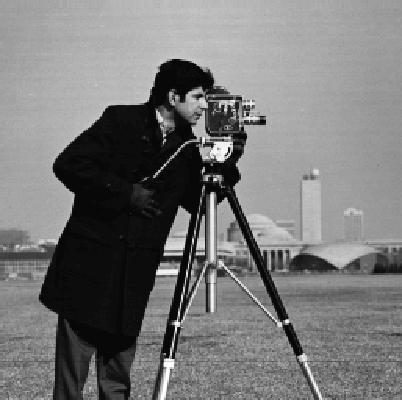} &
			\includegraphics[height = 0.15\textwidth]{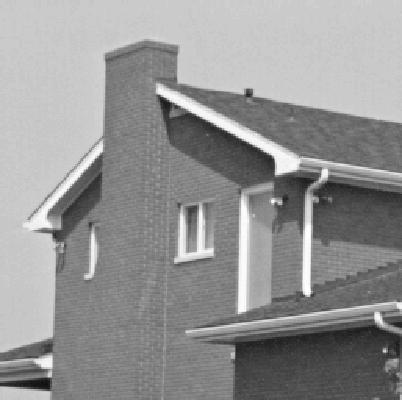} &
			\includegraphics[height = 0.15\textwidth]{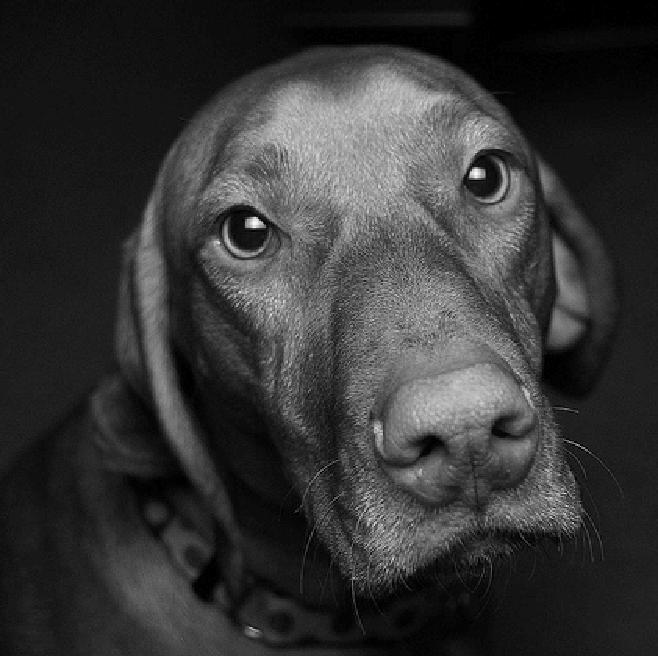} &
			\includegraphics[height = 0.15\textwidth]{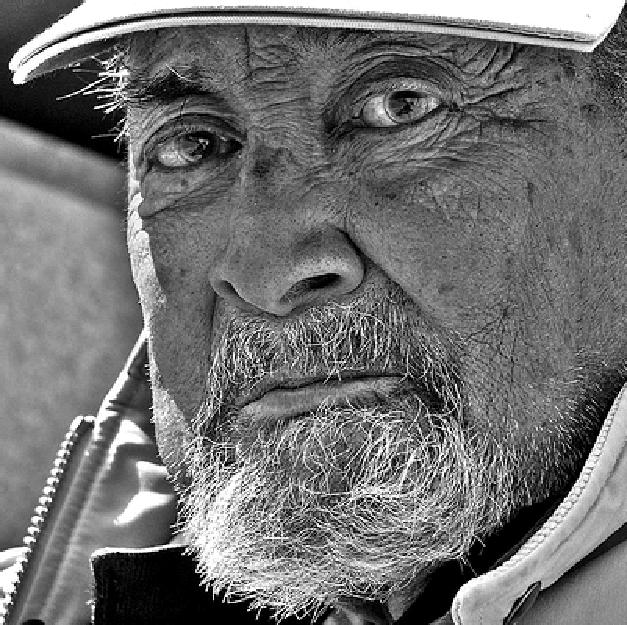} &
			\includegraphics[height = 0.15\textwidth]{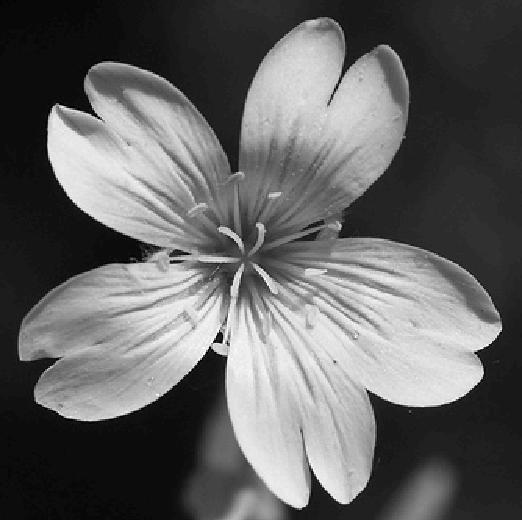} &\\  
			
			\hspace{-1mm}\parbox[b][4em][s]{0.16\textwidth}{Noisy input}&       
			\includegraphics[height = 0.15\textwidth]{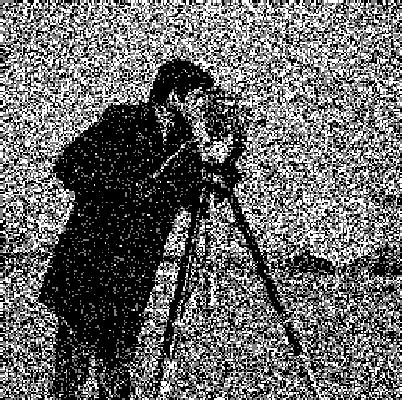} &
			\includegraphics[height = 0.15\textwidth]{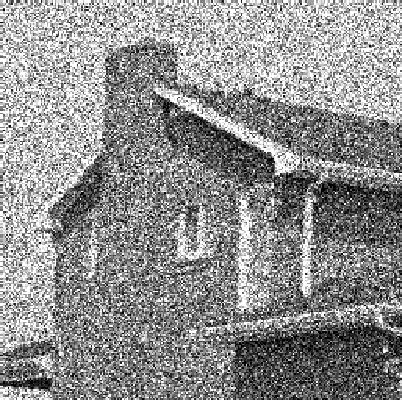} &
			\includegraphics[height = 0.15\textwidth]{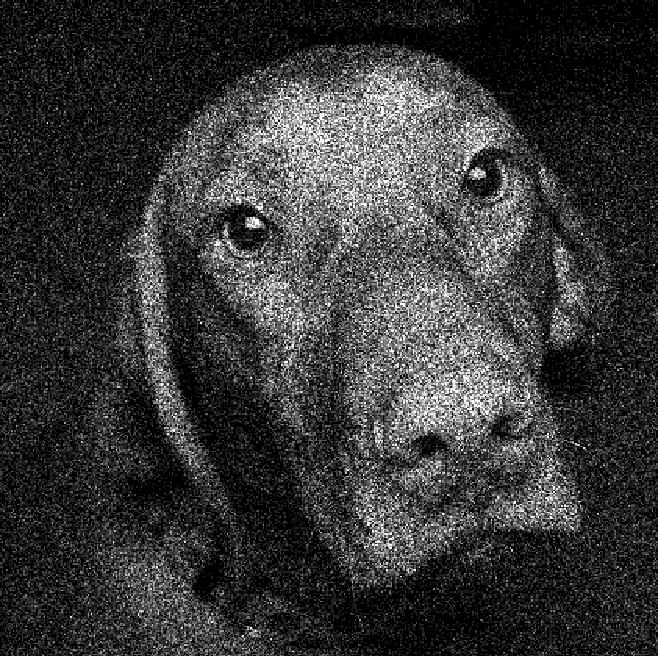} &
			\includegraphics[height = 0.15\textwidth]{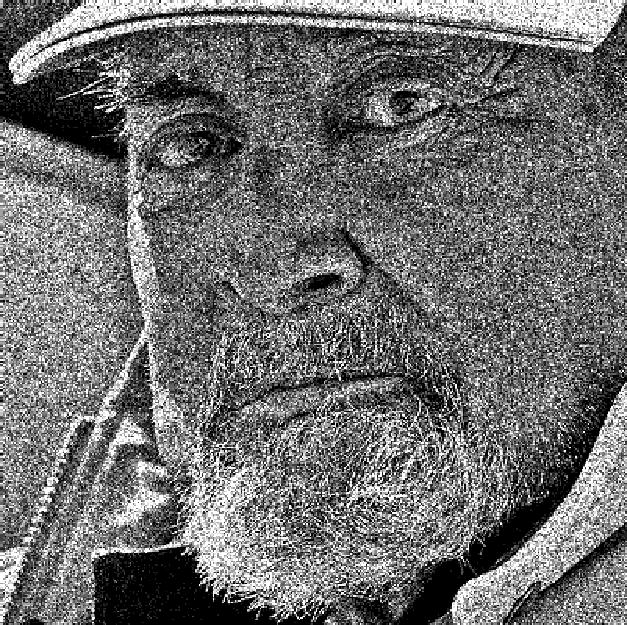} &
			\includegraphics[height = 0.15\textwidth]{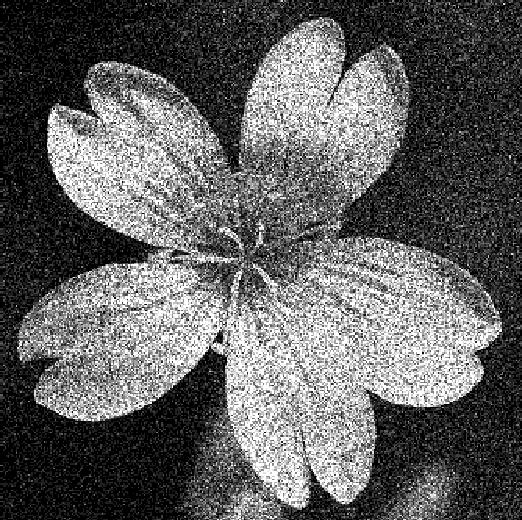} &\\  
			
			\hspace{-1mm}\parbox[b][4em][s]{0.16\textwidth}{Denoised image}&        
			\includegraphics[height = 0.15\textwidth]{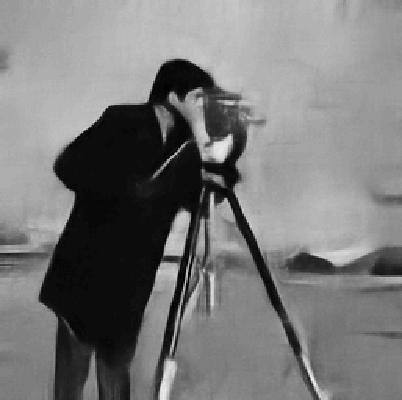} &
			\includegraphics[height = 0.15\textwidth]{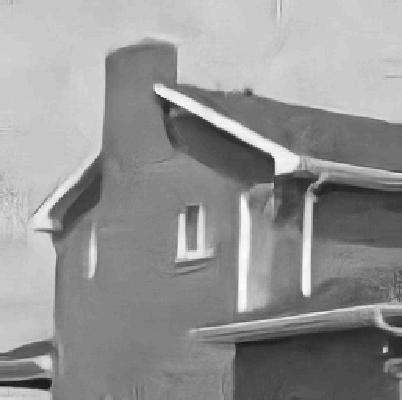} &
			\includegraphics[height = 0.15\textwidth]{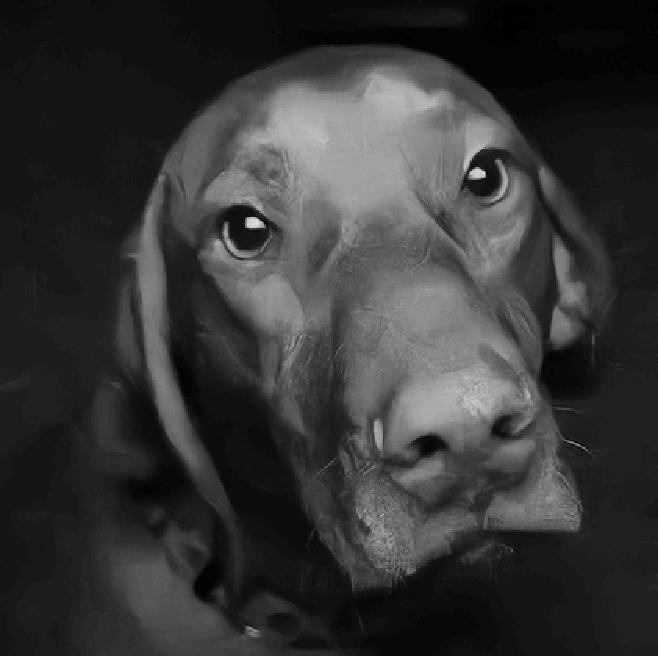} &
			\includegraphics[height = 0.15\textwidth]{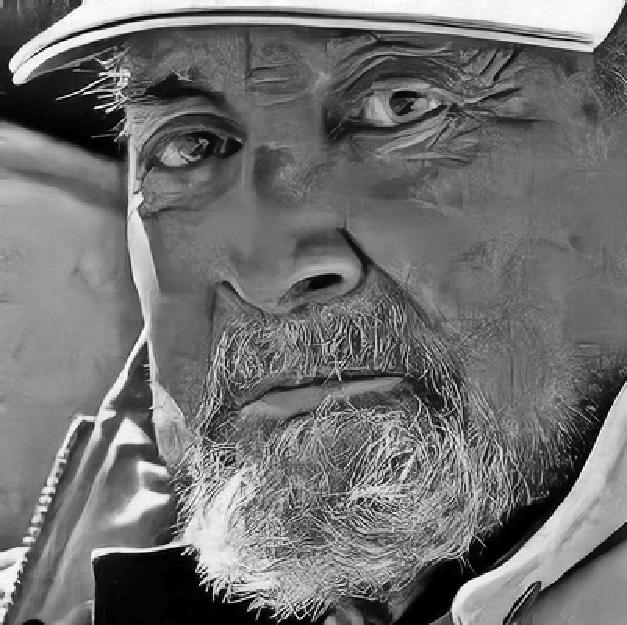} &
			\includegraphics[height = 0.15\textwidth]{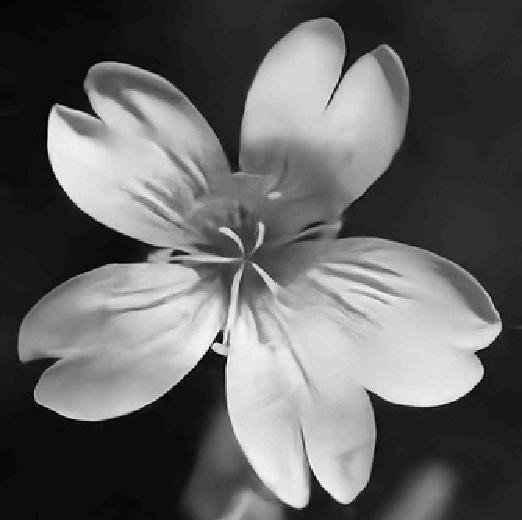} &\\  
			
			\hspace{-1mm}\parbox[b][4em][s]{0.16\textwidth}{Error after 5 layers}&        
			\includegraphics[height = 0.15\textwidth]{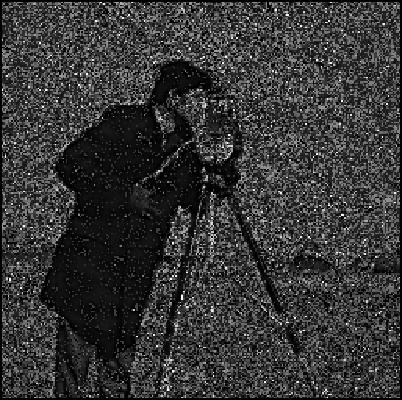} &
			\includegraphics[height = 0.15\textwidth]{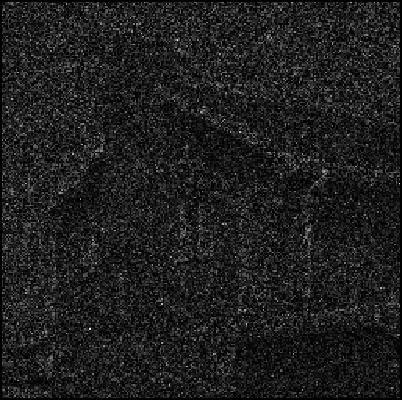} &
			\includegraphics[height = 0.15\textwidth]{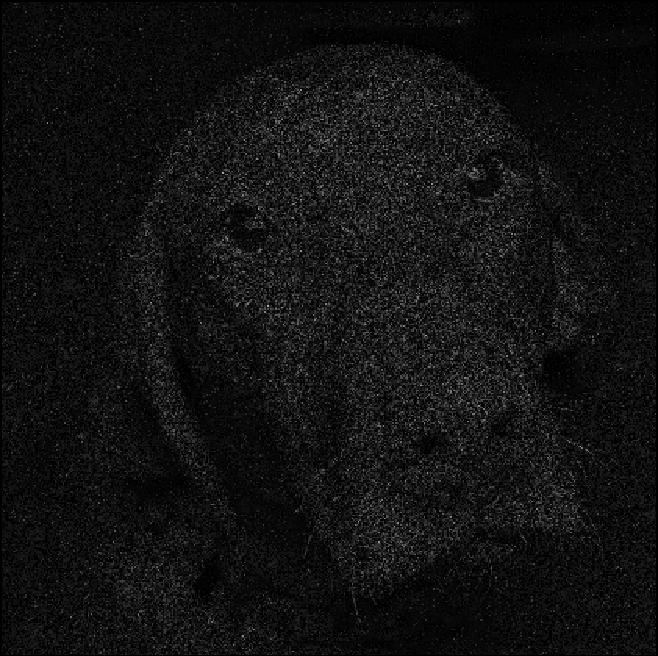} &
			\includegraphics[height = 0.15\textwidth]{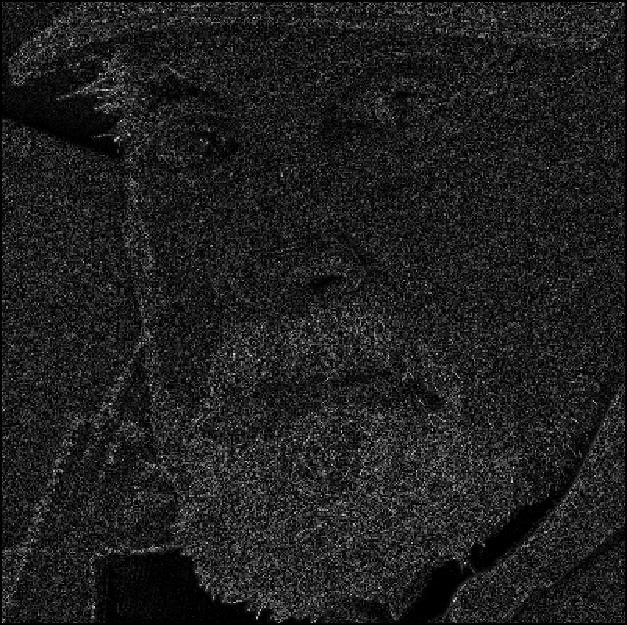} &
			\includegraphics[height = 0.15\textwidth]{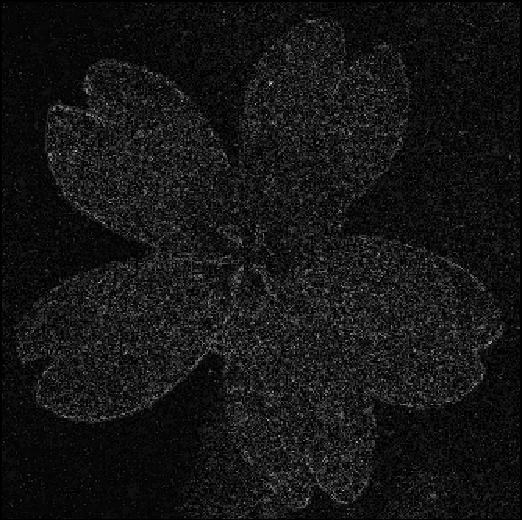} &\\ 
			
			\hspace{-1mm}\parbox[b][4em][s]{0.16\textwidth}{Error after 10 layers}&       
			\includegraphics[height = 0.15\textwidth]{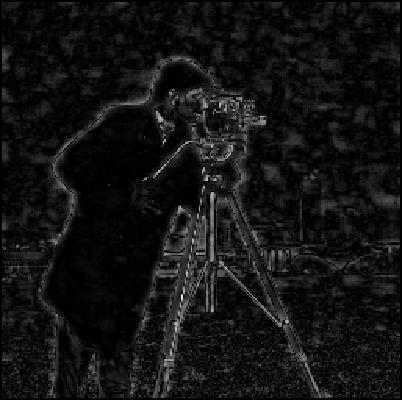} &
			\includegraphics[height = 0.15\textwidth]{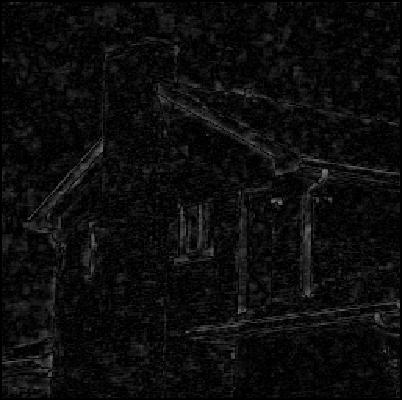} &
			\includegraphics[height = 0.15\textwidth]{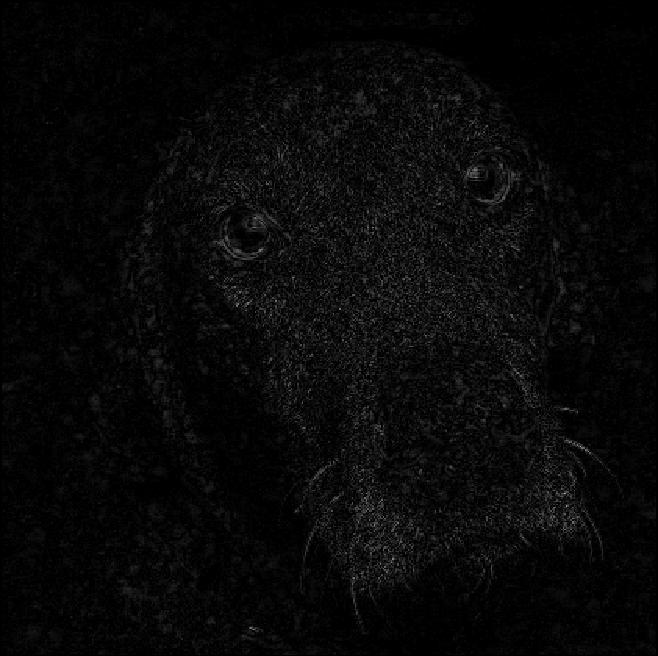} &
			\includegraphics[height = 0.15\textwidth]{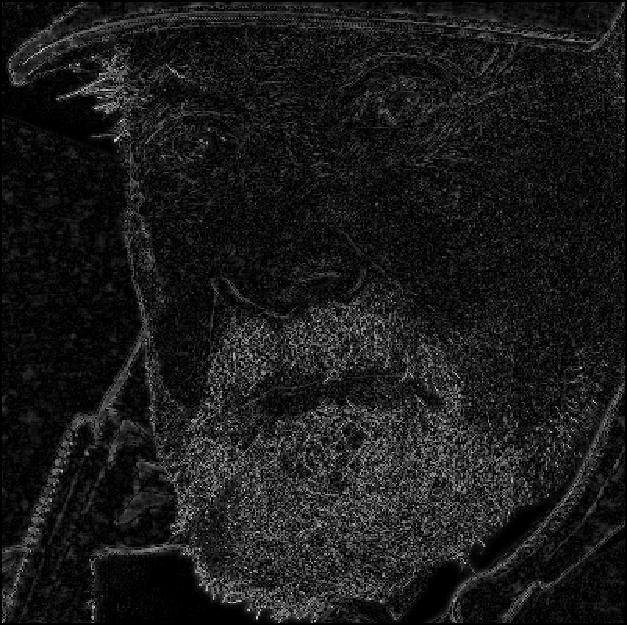} &
			\includegraphics[height = 0.15\textwidth]{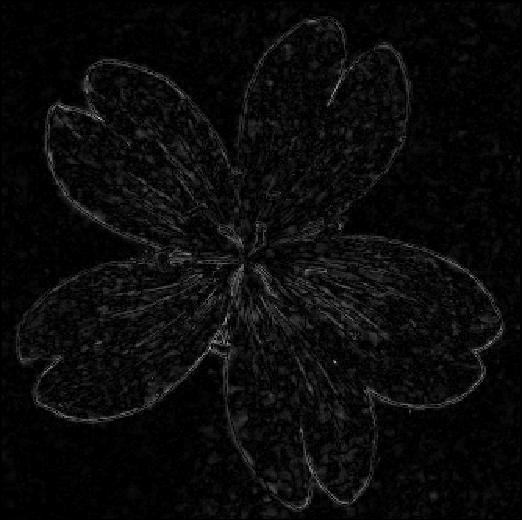} &\\ 
			
			\hspace{-1mm}\parbox[b][4em][s]{0.16\textwidth}{Error after 20 layers (output)}&
			\includegraphics[height = 0.15\textwidth]{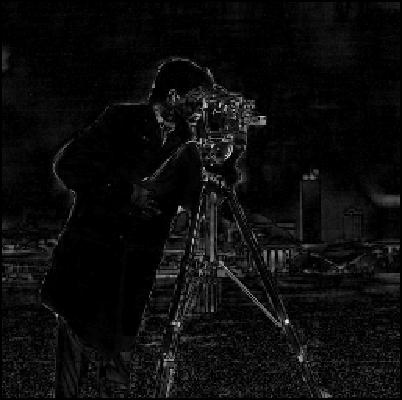} &
			\includegraphics[height = 0.15\textwidth]{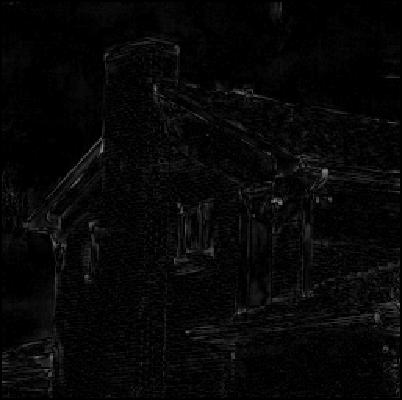} &
			\includegraphics[height = 0.15\textwidth]{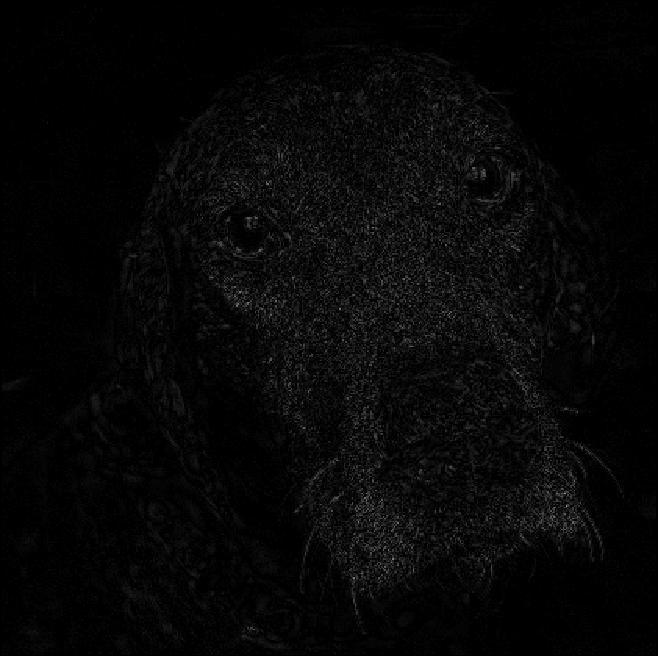} &
			\includegraphics[height = 0.15\textwidth]{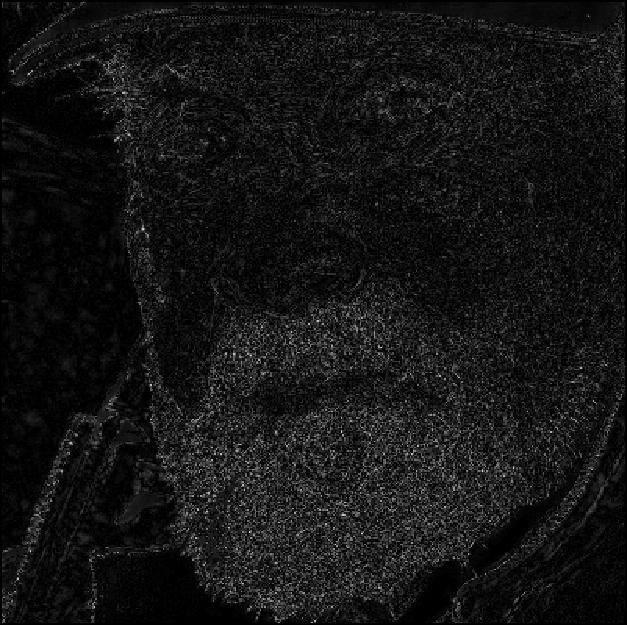} &
			\includegraphics[height = 0.15\textwidth]{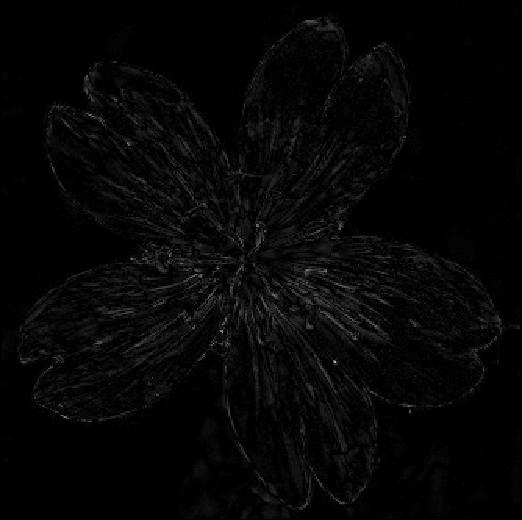} &\\ 
			
			
			\hspace{-1mm}\parbox[b][4em][s]{0.16\textwidth}{RMSE at different layers}&     
			\includegraphics[width = 0.15\textwidth]{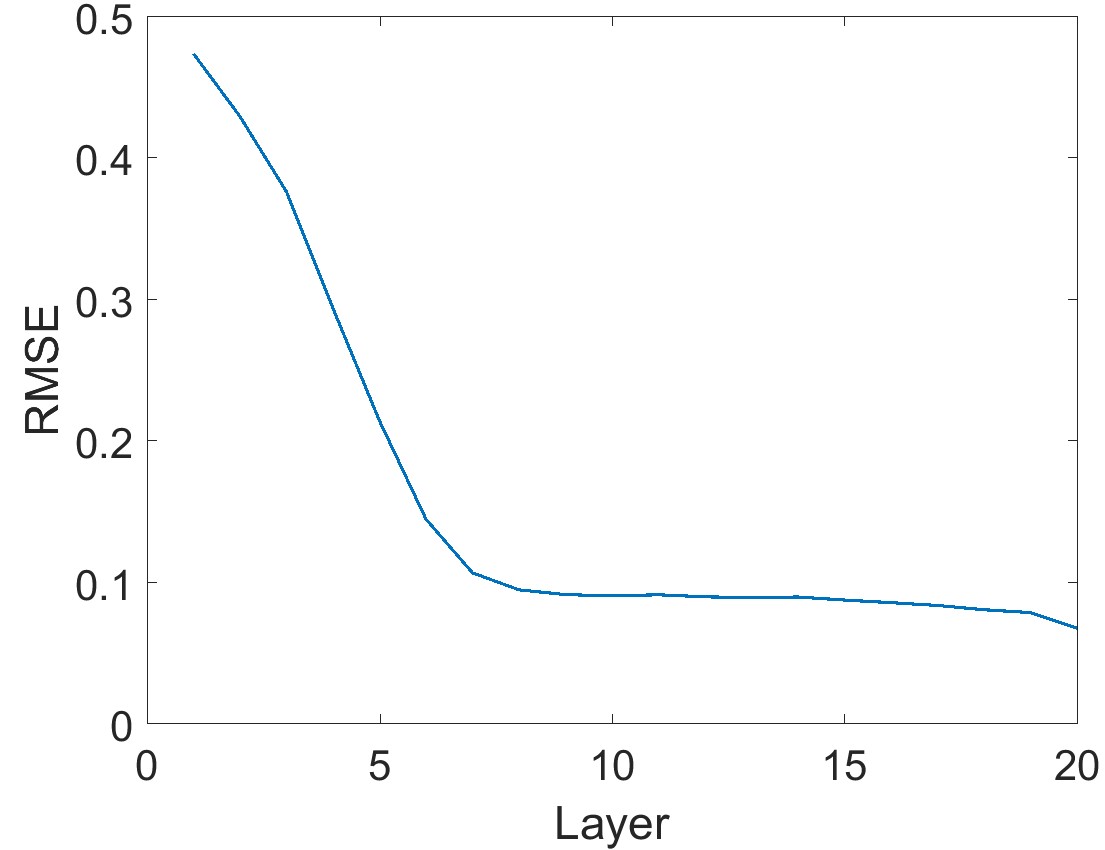}&
			\includegraphics[width = 0.15\textwidth]{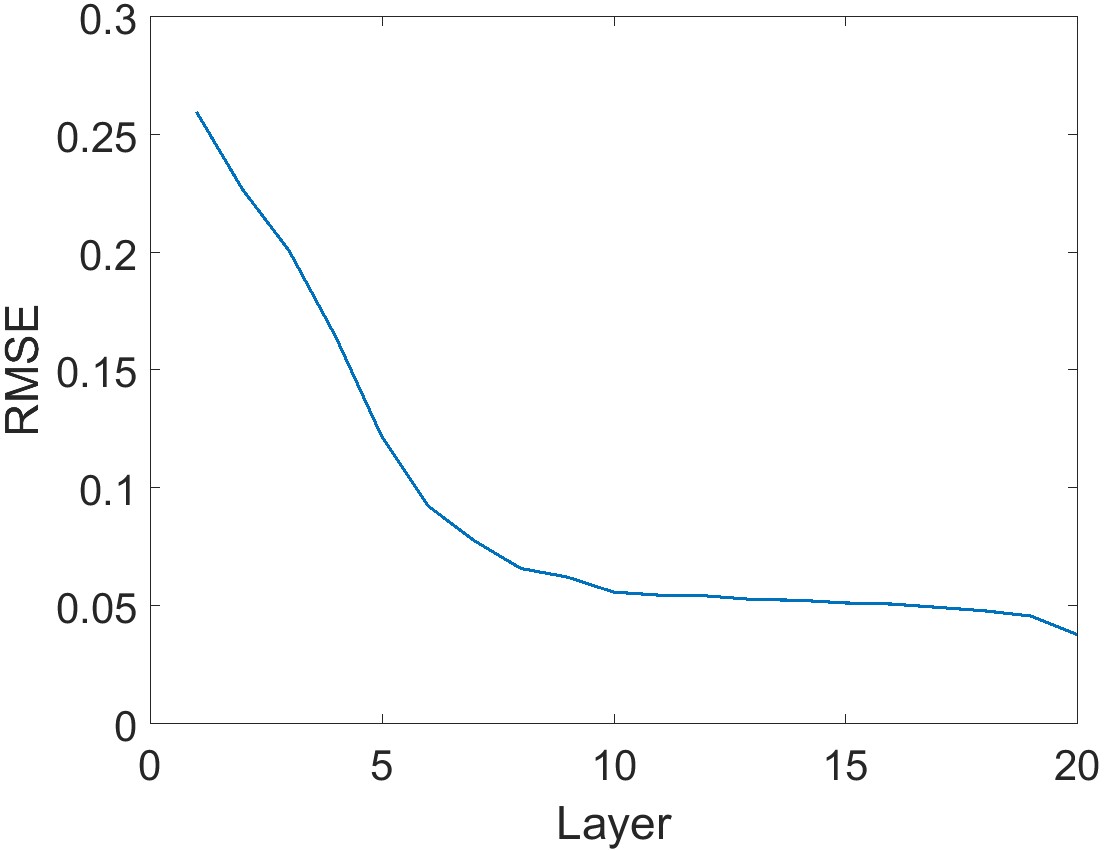}&
			\includegraphics[width = 0.15\textwidth]{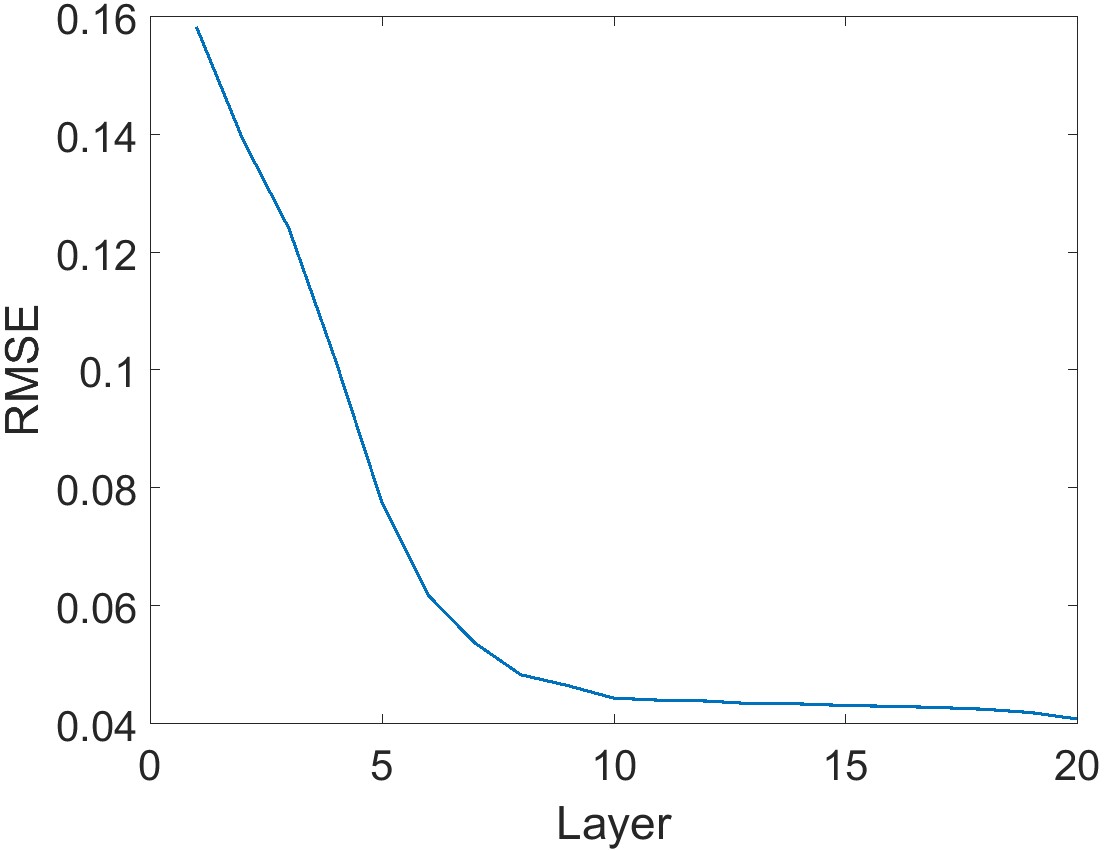}&
			\includegraphics[width = 0.15\textwidth]{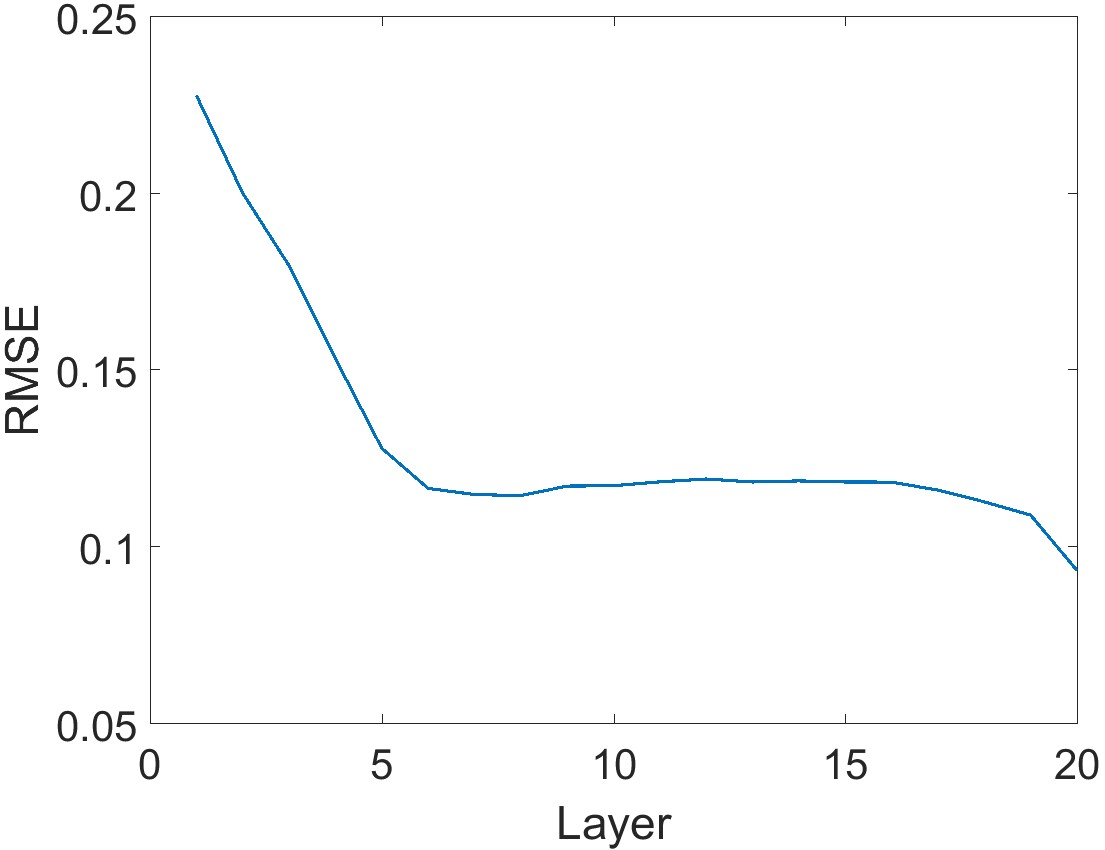}&
			\includegraphics[width = 0.15\textwidth]{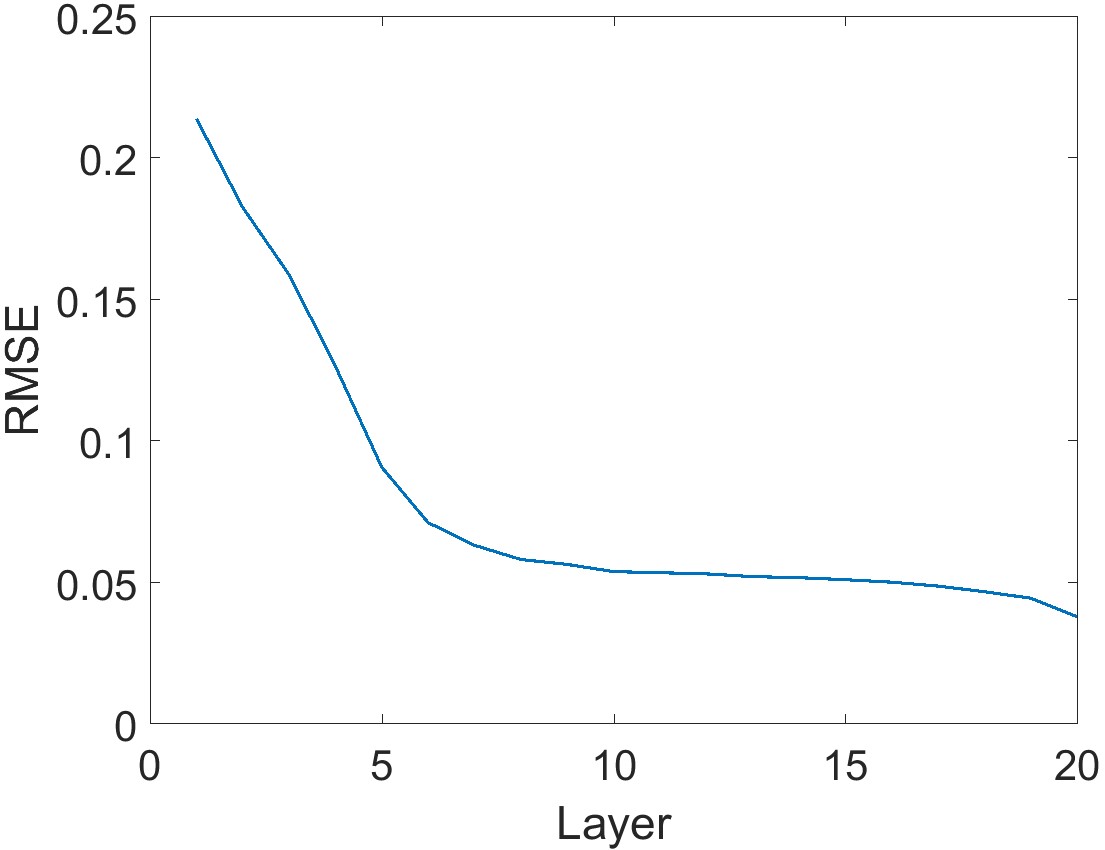}&\\         
			
			\hspace{-1mm}\parbox[b][4em][s]{0.16\textwidth}{Layer contributing the most to each pixel}&
			\includegraphics[height = 0.15\textwidth]{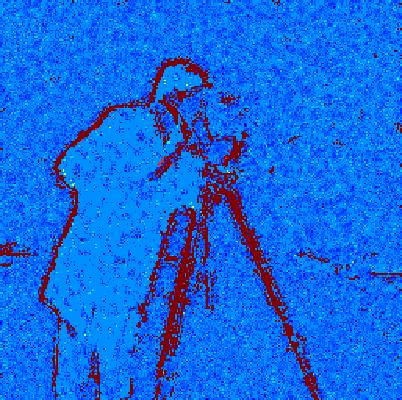} &
			\includegraphics[height = 0.15\textwidth]{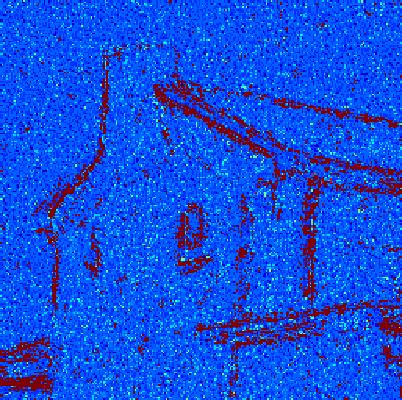} &
			\includegraphics[height = 0.15\textwidth]{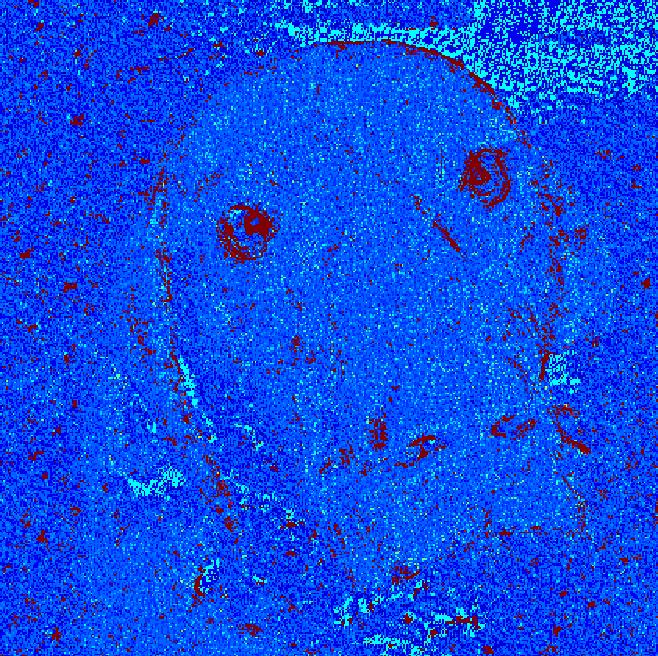} &
			\includegraphics[height = 0.15\textwidth]{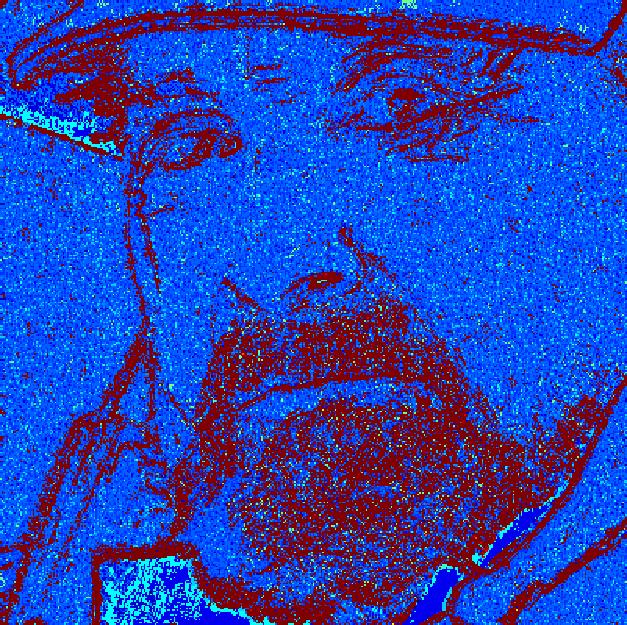} &
			\includegraphics[height = 0.15\textwidth]{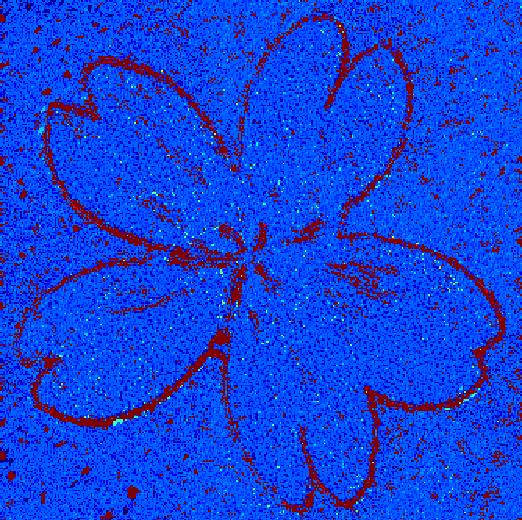} &
			\hspace{-1.5mm} \includegraphics[width = 0.018\textwidth]{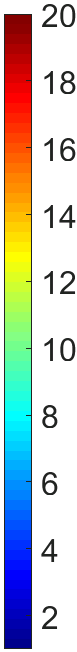} \\

		\end{tabular}   \\
	\end{centering}
	\smallskip
	\caption{\small \textbf{Gradual denoising process. } Images are best viewed electronically. The reader is encouraged to zoom in for a better view. Please refer to the text for more details.}
	\label{fig_layer_selection}
\end{figure*}

\begin{figure*}[]
	\centering    
	\begin{tabular}{c@{\hskip 0.005\textwidth}c@{\hskip 0.005\textwidth}c@{\hskip 0.005\textwidth}c}    
		Ground Truth & Noisy & I+VST+BM3D \cite{Azzari16Variance} & DenoiseNet \\
		
		\includegraphics[width = 0.23\textwidth]{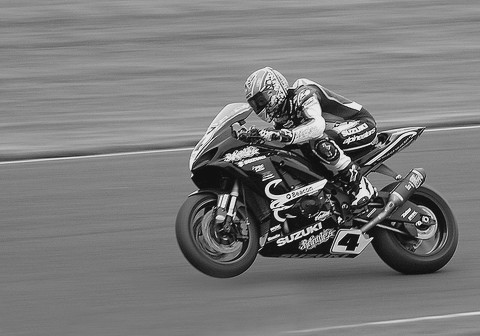} &
		\includegraphics[width = 0.23\textwidth]{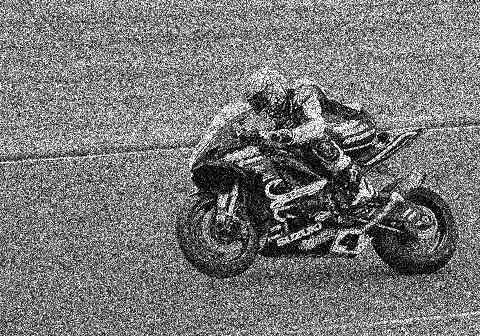} &
		\includegraphics[width = 0.23\textwidth]{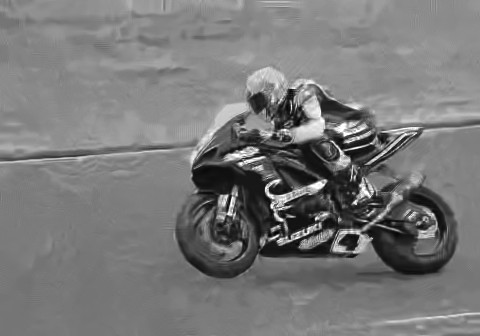} &
		\includegraphics[width = 0.23\textwidth]{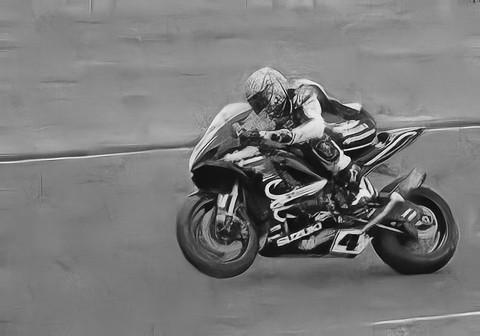}  \\ 
		&\medskip  & $24.56$ dB & $25.83$ dB \\          
		
		\includegraphics[width = 0.23\textwidth]{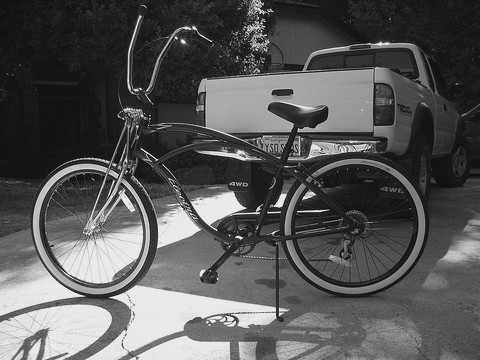} &
		\includegraphics[width = 0.23\textwidth]{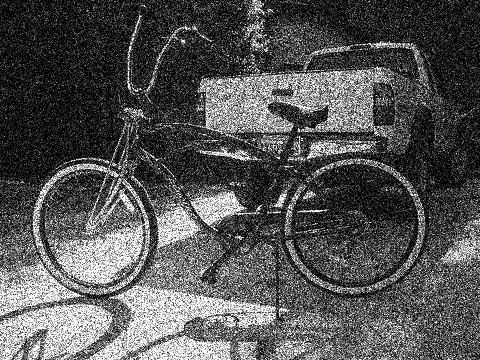} &
		\includegraphics[width = 0.23\textwidth]{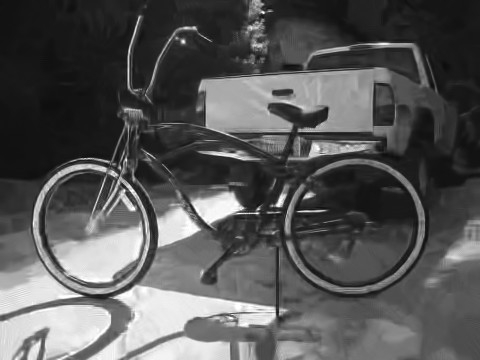} &
		\includegraphics[width = 0.23\textwidth]{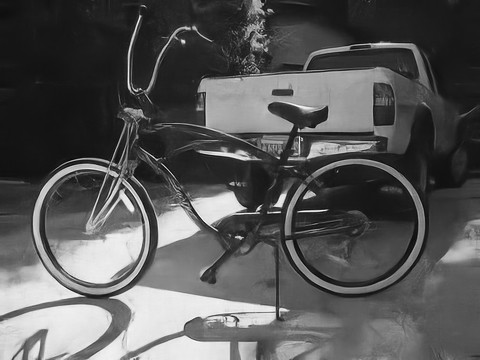}  \\ 
		&\medskip  & $24.76$ dB & $26.00$ dB \\ 
		
		\includegraphics[width = 0.23\textwidth]{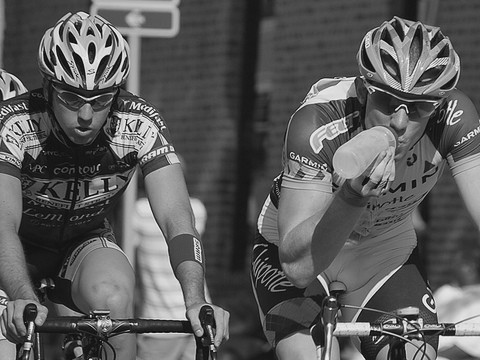} &
		\includegraphics[width = 0.23\textwidth]{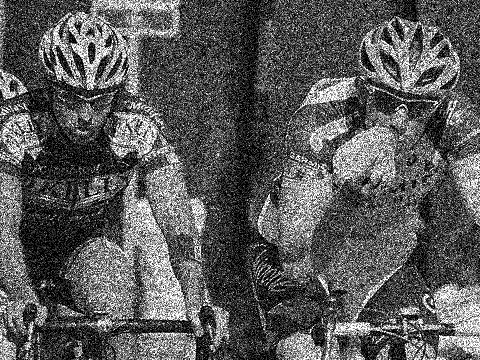} &
		\includegraphics[width = 0.23\textwidth]{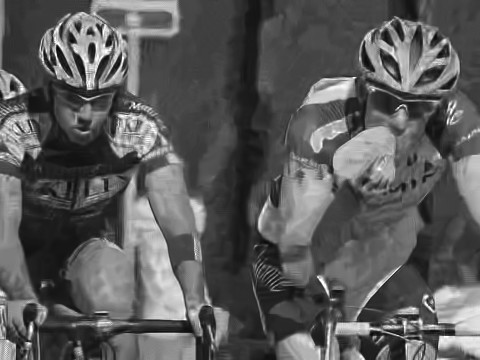} &
		\includegraphics[width = 0.23\textwidth]{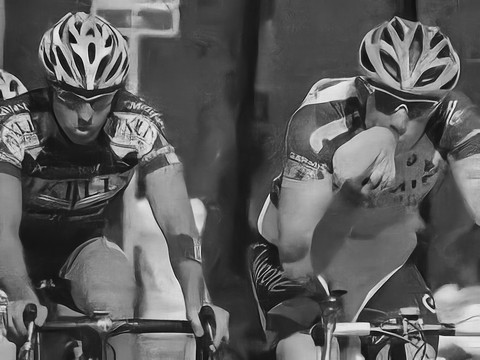}  \\ 
		\medskip &  & $23.40$ dB & $24.66$ dB \\ 
		
		\includegraphics[width = 0.23\textwidth]{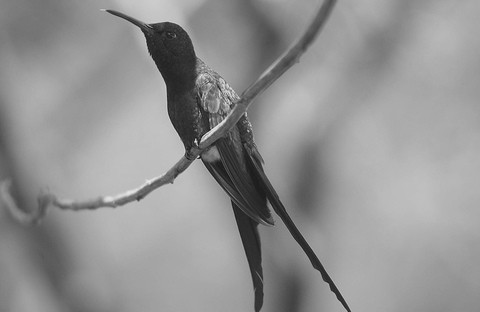} &
		\includegraphics[width = 0.23\textwidth]{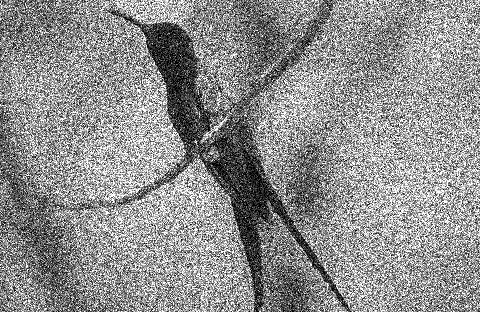} &
		\includegraphics[width = 0.23\textwidth]{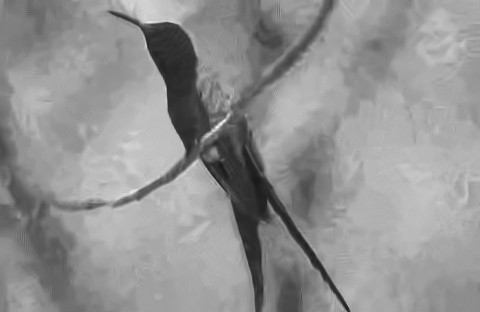} &
		\includegraphics[width = 0.23\textwidth]{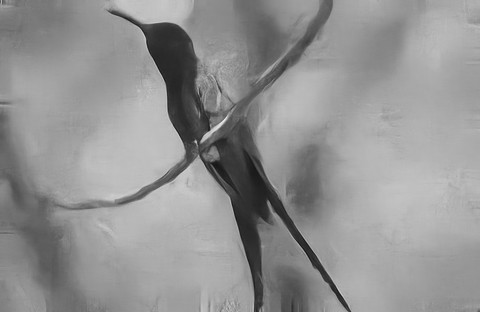}  \\ 
		&  & $30.14$ dB & $31.61$ dB \\

	\end{tabular} 
	\caption{\textbf{Denoising examples from PASCAL VOC 2010.} Presented are images from PASCAL VOC denoised by I+VST+BM3D and our class-agnostic denoiser for peak $8$. PSNR values appear below each of the images. See text for more details.}
	\label{fig_large_pascal}
\end{figure*}

\section{Conclusion}
\label{sec:conc}

In this work we have proposed a CNN-based Poisson image denoiser. Interestingly, our network achieves state-of-the-art performance without explicitly taking into consideration the nature of the noise in hand, but rather it is learned implicitly from the data. We further show how additional knowledge of the image class boosts performance both quantitatively and qualitatively. We believe this offers a flexible learning-based alternative to previous heavily engineered solutions, which is both powerful and fast, and thus may also be adopted to more general types of image enhancement such as Poisson-Gaussian denoising, super-resolution, deblurring, etc.

{\small
\bibliographystyle{ieee}
\bibliography{egbib}
}

\end{document}